%% file: main_preprint.tex
\documentclass[11pt]{article}

\PassOptionsToPackage{numbers,sort&compress}{natbib}

\usepackage[preprint]{acl}

\newif\ifpreprint
\preprinttrue  

\usepackage{times}
\usepackage{latexsym}

\usepackage[T1]{fontenc}
\usepackage[utf8]{inputenc}

\usepackage{microtype}
\usepackage{graphicx}

\usepackage{booktabs}    
\usepackage{multirow}    
\usepackage{makecell}    
\usepackage{xcolor}      
\usepackage{colortbl}    
\usepackage{array}       
\usepackage{tabularx}    
\usepackage{adjustbox}   
\usepackage{arydshln}    

\arrayrulecolor{black}
\usepackage{pifont}      
\usepackage{amsmath}     
\usepackage[most]{tcolorbox}  

\newtcolorbox{paperbox}[1]{
  colback=paperlavenderlight,
  colframe=papernavy,
  coltitle=white,
  fonttitle=\bfseries\scshape\small,
  title=#1,
  arc=1.5mm,
  boxrule=0.5pt,
  left=4pt, right=4pt, top=3pt, bottom=3pt,
  breakable,
  before skip=4pt, after skip=4pt
}
\usepackage{amssymb}     
\usepackage{enumitem}    
\usepackage{textcomp}    
\usepackage{tikz}        
\usepackage{placeins}    
\usetikzlibrary{positioning, arrows.meta, shapes.geometric, calc, fit, backgrounds}

\newcommand{\cmark}{\ding{51}}
\newcommand{\xmark}{\ding{55}}
\newcommand{\dash}{\textendash}

\definecolor{paperbeige}{HTML}{E5C8A0}
\definecolor{paperbeigelight}{HTML}{F1E1C9}
\definecolor{paperlavender}{HTML}{9193B4}
\definecolor{paperlavenderlight}{HTML}{CDCFE1}
\definecolor{paperpink}{HTML}{BD9AAD}
\definecolor{paperpinklight}{HTML}{EAD4D5}
\definecolor{papernavy}{HTML}{2F2D54}
\definecolor{papernavylight}{HTML}{5C6592}

\definecolor{rowbeige}{HTML}{F1E1C9}
\definecolor{rowlavender}{HTML}{CDCFE1}
\definecolor{rowpink}{HTML}{EAD4D5}
\definecolor{rowdiagnostic}{HTML}{E0DCEC}
\definecolor{rowhighlight}{HTML}{C9B894}

\newcommand{\strong}{\textcolor{papernavy}{\cmark\cmark\cmark}}
\newcommand{\medium}{\textcolor{paperlavender}{\cmark\cmark}}
\newcommand{\weak}{\textcolor{paperpink}{\cmark}}
\renewcommand{\xmark}{\textcolor{paperpink}{\ding{55}}}
\renewcommand{\cmark}{\textcolor{papernavy}{\ding{51}}}

\newcommand{\mentalmap}{\textsc{MentalMap}}

\usepackage{caption}
\title{Do LLMs Build World Models From Text? \\ A Multilingual Diagnostic of Spatial Reasoning}

\author{
  Zhikai Pan$^{1*}$ \quad
  Chih-Ting Liao$^{1*\dagger\ddagger}$ \quad
  Chunrui Liu$^{2}$ \quad
  Xi Xiao$^{3}$ \\[2pt]
  \textbf{Yitong Qiao$^{4}$} \quad
  \textbf{Chunlei Meng$^{5}$} \quad
  \textbf{Zhangquan Chen$^{6}$} \quad
  \textbf{Xin Cao$^{1}$} \\[6pt]
  $^{1}$University of New South Wales \quad
  $^{2}$Essential Energy \quad
  $^{3}$University of Alabama at Birmingham \\
  $^{4}$Zhejiang University \quad
  $^{5}$Fudan University \quad
  $^{6}$Tsinghua University \\[4pt]
  \texttt{mill.liao@unsw.edu.au} \\[2pt]
  \small{$^{*}$Equal contribution. \quad $^{\dagger}$Corresponding author. \quad $^{\ddagger}$Project lead.}
}

\begin{document}
\maketitle

\input{sections/00_abstract}

\input{sections/01_introduction}
\input{sections/02_related_work}
\input{sections/03_benchmark_design}
\input{sections/04_experiments_and_results}
\input{sections/07_conclusion}

\newpage
\input{sections/08_limitations}

\bibliography{custom}

\appendix
\input{sections/A_appendix}

\end{document}

%% file: sections/00_abstract.tex
\begin{abstract}
Whether large language models (LLMs) construct internal spatial world models from pure-text descriptions---or merely retrieve memorised co-occurrences---remains contested, and whether any such capability transfers across typologically diverse languages has not been systematically probed. We answer through \textbf{\mentalmap{}}, a multilingual diagnostic benchmark with a two-axis design. The \emph{capability axis} is a six-level staircase (L0--L5) progressing from atomic spatial facts to generative world-graph output. The \emph{diagnostic axis} comprises four orthogonal lenses---frame of reference, reading-direction bias, reasoning-effort allocation, and hallucination---each instrumented within the staircase. \mentalmap{} is grounded in 100 ProcTHOR \citep{deitke2022procthor} household scenes, spans eight typologically diverse languages plus a structured-text control, and comprises 39 task families across 1{,}950 evaluation cells (over 47{,}000 scored items per model, half a million across the evaluated panel). To our knowledge, it is the first benchmark to combine pure-text spatial reasoning, multilingual coverage, and generative world-graph evaluation in a single coherent design. Evaluating thirteen LLMs across frontier closed-source, open-weight 7--10B families, and two mid-scale open-weight scale controls (27--32B), we surface seven findings that together argue against treating spatial competence as a single capability. \textbf{(i)} The capability staircase is non-monotonic: a universal cliff at L3 viewpoint reasoning marks a discrete transition from static comprehension (recall) to active spatial reasoning (construction)---across all evaluated (model, language) cells in the cliff-diagnostic regime (L0$>$40\%), none reaches half of its L0 atomic performance at L3. \textbf{(ii)} The cliff is sub-task structured, dissociating frame-transformation collapse from boolean consistency-probe guessing. \textbf{(iii)} Chain-of-thought prompting is not a universal booster, helping DeepSeek-V4-Flash by $+32$pp at L3 but hurting Qwen2.5-7B by $-16$pp at the same level and disrupting strict JSON output for most open-weight models at L5. \textbf{(iv)} At generative world-graph output, node identification (object existence) and edge extraction (containment relations) dissociate sharply, and strict pass rate and partial-credit graph F1 give materially different model rankings. \textbf{(v)} The L3 cliff persists under two mid-scale open-weight scale controls (27--32B), while open-weight 7--10B models close the closed-source gap specifically at L5---together indicating fine-tuning regime rather than parameter scale drives structured-output competence. \textbf{(vi)} Hierarchical clustering of per-model multilingual profiles recovers the Latin / CJK / (Arabic, Thai) script typology from model-performance correlations alone, and separates models into narrow- and broad-profile regimes that cut across scale. \textbf{(vii)} A cross-lingual human pilot under the identical pure-text protocol reproduces the cliff in every evaluated language---L3$=$0\% in all eight languages while L4 is remarkably language-invariant ($\sim$41\%, std 3.3pp)---direct evidence that the cliff reflects a working-memory bottleneck of the text modality rather than an LLM-specific or language-specific deficit. The combined findings reframe pure-text spatial reasoning as a multi-axis world-modeling problem rather than a single-number leaderboard, and motivate scratchpad-augmented and multimodal protocols as the natural next probe. \mentalmap{}, including all data, schema validators, and evaluation harness, will be publicly released.
\end{abstract}

%% file: sections/01_introduction.tex
\section{Introduction}
\label{sec:introduction}

\emph{Do large language models (LLMs) build internal world models from pure-text descriptions of physical scenes, and if so, are these models robust across languages?} The question matters because spatial reasoning---constructing, manipulating, and updating representations of physical configurations from linguistic input---is a foundational pillar of grounded intelligence \citep{levinson2003space,talmy2000toward}, and because the answer determines whether LLMs can be trusted as the language interface for embodied agents, robotics planners, and grounded dialogue systems.

Existing spatial-reasoning evaluations leave the question unanswered. Three concurrent gaps obscure what LLMs actually understand about space: \textbf{(i) English monolingualism}---languages differ systematically in their preferred frames of reference \citep{levinson2003space,pederson1998semantic}, yet LLM spatial benchmarks remain almost exclusively English \citep{premsri2025forest,choukrani2025babybench,sit-bench-2026}; \textbf{(ii) single-dimension evaluation}---benchmarks probe perspective \citep{spatialtext2026}, state \citep{choukrani2025babybench}, or graph generation \citep{yang2025tsg} in isolation, obscuring their interactions; and \textbf{(iii) surface-versus-substance ambiguity}---high accuracy may reflect statistical co-occurrences rather than coherent internal world models \citep{spatialtext2026,probing-primitives-2026}. Diagnostic instruments from cognitive linguistics \citep{vanderhenst2008influence} and long-reasoning analysis \citep{wang2025underthinking,optimal-thinking-bench-2025} have not been systematically applied to spatial reasoning.

\noindent\textbf{Our contribution.} We present \textbf{\mentalmap{}}, a pure-text spatial-reasoning benchmark designed to close all three gaps through a two-tier architecture. The \emph{first axis} is a six-level capability staircase grounded in ProcTHOR \citep{deitke2022procthor} household scenes, structured as two principled tiers: \emph{static comprehension} (L0--L2, retrieving stated spatial facts) and \emph{active spatial reasoning} (L3--L5, transforming or generating new spatial configurations, culminating in generative world-graph output). The two-tier division is motivated by the recall--construction distinction in mental-rotation and egocentric--allocentric studies \citep{shepard1971mental,burgess2006spatial}---and, as we show, empirically validated by a universal performance cliff at the tier boundary. The \emph{second axis} is a four-lens diagnostic dimension orthogonal to the staircase: \emph{frame of reference}, \emph{reading-direction bias}, \emph{reasoning effort}, and \emph{hallucination}. \mentalmap{} spans eight typologically diverse languages---English, Mandarin, Japanese, Korean, Spanish, Arabic, Thai, German---covering all three Levinsonian frame classes and both LTR and RTL writing systems, plus a structured-text control that isolates spatial reasoning from language quality. To our knowledge, \mentalmap{} is the first benchmark to combine pure-text spatial reasoning, multilingual coverage, and generative world-graph evaluation in a single coherent design (Table~\ref{tab:benchmark_comparison}).

\noindent\textbf{Headline findings.} Evaluating thirteen LLMs across frontier closed-source, open-weight 7--10B families, scale controls (Qwen2.5-32B, Gemma-3-27B), and a vision-language ablation (Qwen2.5-VL), we report seven findings: \textbf{(F1)} a universal L3 cliff---the staircase \emph{breaks} at the static-to-active boundary, with every (model, language) cell in the cliff-diagnostic regime scoring below half of its L0 atomic performance at L3 viewpoint reasoning; \textbf{(F2)} the cliff is sub-task structured, dissociating frame-transformation collapse from consistency-probe guessing; \textbf{(F3)} chain-of-thought is not a universal booster, helping DeepSeek by $+32$pp at L3 but hurting Qwen2.5-7B by $-16$pp at the same level; \textbf{(F4)} node identification and relation extraction dissociate sharply at L5, and strict pass rate and partial-credit graph F1 give materially different model rankings; \textbf{(F5)} the L3 cliff persists under two mid-scale open-weight scale controls (27--32B) while open-weight L5 catches up to closed-source, indicating fine-tuning regime drives the structured-output convergence; \textbf{(F6)} hierarchical clustering of per-model multilingual profiles recovers the Latin / CJK / (Arabic, Thai) script typology from performance correlations alone, separating models into narrow- and broad-profile regimes; \textbf{(F7)} a cross-lingual human pilot under the same pure-text protocol reproduces the L3 cliff in every evaluated language (L3$=$0\% all 8 languages, L4 language-invariant at $\sim$41\%), direct evidence that the cliff reflects a working-memory bottleneck of the text modality. Together, these reframe pure-text spatial reasoning as a multi-axis problem in which the static-to-active boundary marks a discrete, modality-related failure mode rather than a single scalar capability.

\noindent The cliff in F1 should not be read as a claim that L3 is easy for humans: mental rotation and perspective-taking are themselves effortful tasks \citep{shepard1971mental,kozhevnikov2001two}. F7 makes this concrete---under the same pure-text protocol native speakers also collapse on hard L3 items.

%% file: sections/02_related_work.tex
\section{Related Work}
\label{sec:related_work}

\input{tables/benchmark_comparison}

We position \mentalmap{} against five families of prior work (Table~\ref{tab:benchmark_comparison}).

\noindent\textbf{Pure-text spatial reasoning.} StepGame \citep{shi2022stepgame} and SpaRC/SpaRP \citep{rizvi2024sparc} probe atomic spatial-relation reasoning on synthetic grids; SpatialText \citep{spatialtext2026}, SiT-Bench \citep{sit-bench-2026}, LLM-BabyBench \citep{choukrani2025babybench}, and MazeEval \citep{mazeeval2025} add scene grounding and state tracking. Closest in framing are recent grid-world probes \citep{li2026maze,martorell2025navigation} that ask whether LLMs build spatial world models from text; both find that performance depends sharply on representation format and prompting regime, consistent with our F3 and F5 findings. \mentalmap{} differs from these by (i) eight-language multilingual coverage spanning all Levinsonian frame classes, (ii) ProcTHOR household scenes rather than abstract grids, and (iii) a two-axis design with both a capability staircase and orthogonal diagnostic lenses. A complementary line probes whether LLMs encode coherent latent world representations rather than testing behavioural outputs \citep{li2024probing,hao2023reasoning,vafa2024evaluating}; \mentalmap{} is behavioural and orthogonal. All prior text-only spatial work shares three limitations: 2D abstract grids or low-diversity worlds, at most two languages, and no diagnostic axes that distinguish surface pattern-matching from genuine spatial reasoning.

\noindent\textbf{Frame-of-reference evaluation.} Cognitive linguistics distinguishes relative, absolute, and intrinsic frames \citep{levinson2003space,majid2004can,tenbrink2011reference}. FoREST \citep{premsri2025forest} operationalizes this axis but is English-only; COMFORT \citep{zhang2024comfort} extends to multilingual but is VLM-based on synthetic 3D images.

\noindent\textbf{Structured-output and world-state benchmarks.} TSG Bench \citep{yang2025tsg}, EAI Transition \citep{li2024embodied}, and FloorplanQA \citep{floorplanqa2026} test LLM-emitted structured world representations; ByteSized32-State-Prediction \citep{wang2024bytesized32} and TEXT2WORLD \citep{hu2025text2world} evaluate text-based simulators and symbolic world-model generation. \mentalmap{} differs by being pure-text, coupling structured graph output with a static-to-active capability staircase under multilingual coverage, and reporting per-level cell-resolved metrics rather than aggregate executability.

\noindent\textbf{Multimodal world-model benchmarks.} \ifpreprint SpaMEM~\citep{liao2026spamem} probes dynamic spatial memory for VLMs in embodied settings; \mentalmap{} isolates the pure-text axis that such multimodal benchmarks bundle with vision.\fi

\noindent\textbf{Diagnostic-lens benchmarks.} OptimalThinkingBench \citep{optimal-thinking-bench-2025} and Sys2Bench \citep{sys2bench2025} examine reasoning-effort allocation on mathematics and logic, not spatial; general hallucination benchmarks \citep{faithjudge2025,hhem2025} target factual summarization, not the spatial-domain failure modes of phantom receptacles or node--edge dissociation.

%% file: tables/benchmark_comparison.tex
\begin{table*}[t]
\centering
\scriptsize
\setlength{\tabcolsep}{2pt}
\renewcommand{\arraystretch}{1.05}

\begin{adjustbox}{center, max width=\textwidth}
\begin{tabular}{
p{3.2cm}
*{11}{>{\centering\arraybackslash}p{0.78cm}}
}
\toprule
\multirow{2}{*}{\textbf{Benchmark}}
  & \multicolumn{11}{c}{\textbf{Capability dimensions}} \\
  \cmidrule(lr){2-12}
  & \textbf{Text} & \textbf{Sim.}    & \textbf{ML}      & \textbf{Stair.}  & \textbf{Dyn.}   & \textbf{Graph}  & \textbf{View}  & \textbf{FoR}   & \textbf{RTL}  & \textbf{Think.} & \textbf{Halluc.} \\
\midrule

\rowcolor{rowbeige}
\multicolumn{12}{l}{\textbf{\textit{(A) Pure-text spatial reasoning benchmarks}}} \\

StepGame~\citep{shi2022stepgame}                  & \cmark & \xmark & \xmark & \xmark & \xmark & \xmark & \xmark & \xmark & \xmark & \xmark & \xmark \\
SpaRC/SpaRP~\citep{rizvi2024sparc}                & \cmark & \xmark & \xmark & \weak  & \xmark & \xmark & \xmark & \xmark & \xmark & \xmark & \xmark \\
SpatialText~\citep{spatialtext2026}               & \cmark & \cmark & \xmark & \medium & \weak & \xmark & \cmark & \weak  & \xmark & \xmark & \xmark \\
SiT-Bench~\citep{sit-bench-2026}                  & \cmark & \cmark & \xmark & \medium & \cmark & \weak  & \cmark & \xmark & \xmark & \xmark & \xmark \\
LLM-BabyBench~\citep{choukrani2025babybench}      & \cmark & \cmark$^{\dag}$ & \xmark & \weak & \cmark & \weak & \xmark & \xmark & \xmark & \xmark & \xmark \\
MazeEval~\citep{mazeeval2025}                     & \cmark & \cmark$^{\dag}$ & \weak  & \xmark & \cmark & \xmark & \xmark & \xmark & \xmark & \xmark & \xmark \\

\rowcolor{rowpink}
\multicolumn{12}{l}{\textbf{\textit{(B) Frame-of-reference benchmarks}}} \\

FoREST~\citep{premsri2025forest}                  & \cmark & \cmark$^{\dag}$ & \xmark & \xmark & \xmark & \xmark & \cmark & \strong & \xmark & \xmark & \xmark \\
COMFORT~\citep{zhang2024comfort}                  & \xmark$^{\ddag}$ & \cmark$^{\dag}$ & \strong & \xmark & \xmark & \xmark & \cmark & \strong & \xmark & \xmark & \xmark \\

\rowcolor{rowlavender}
\multicolumn{12}{l}{\textbf{\textit{(C) World-state and structured-output benchmarks}}} \\

TSG Bench~\citep{yang2025tsg}                     & \cmark & \xmark & \xmark & \xmark & \xmark & \cmark & \xmark & \xmark & \xmark & \xmark & \xmark \\
EAI Transition~\citep{li2024embodied}             & \cmark & \cmark & \xmark & \xmark & \cmark & \cmark & \xmark & \xmark & \xmark & \xmark & \xmark \\
FloorplanQA~\citep{floorplanqa2026}               & \cmark & \cmark & \xmark & \xmark & \xmark & \weak  & \xmark & \xmark & \xmark & \xmark & \xmark \\
ByteSized32-SP~\citep{wang2024bytesized32}        & \cmark & \cmark & \xmark & \xmark & \cmark & \weak  & \xmark & \xmark & \xmark & \xmark & \xmark \\
TEXT2WORLD~\citep{hu2025text2world}               & \cmark & \cmark & \xmark & \xmark & \cmark & \cmark & \xmark & \xmark & \xmark & \xmark & \xmark \\

\ifpreprint
\rowcolor{rowdiagnostic}
\multicolumn{12}{l}{\textbf{\textit{(D) Multimodal world-model benchmarks (not pure-text)}}} \\

SpaMEM~\citep{liao2026spamem}                     & \xmark$^{\ddag}$ & \cmark & \xmark & \weak  & \cmark & \weak  & \cmark & \xmark & \xmark & \xmark & \cmark \\
\fi

\rowcolor{rowdiagnostic}
\multicolumn{12}{l}{\textbf{\textit{(E) Diagnostic-lens benchmarks (general, not spatial)}}} \\

OptimalThinkingBench~\citep{optimal-thinking-bench-2025} & \cmark & \xmark & \xmark & \xmark & \xmark & \xmark & \xmark & \xmark & \xmark & \strong & \xmark \\
$\alpha$-Law~\citep{alpha-law-2026}               & \cmark & \xmark & \xmark & \xmark & \xmark & \xmark & \xmark & \xmark & \xmark & \xmark & \xmark \\

\midrule
\rowcolor{rowhighlight}
\textbf{\mentalmap{} (Ours)}                           & \strong & \strong & \strong & \strong & \strong & \strong & \strong & \strong & \cmark & \medium$^{\P}$ & \strong \\

\bottomrule
\end{tabular}
\end{adjustbox}

\caption{
\label{tab:benchmark_comparison}
\textbf{\mentalmap{} vs.\ prior benchmarks across eleven capability dimensions.}
Columns: \textbf{Text} pure-text input; \textbf{Sim.}\ simulator-grounded;
\textbf{ML} multilingual; \textbf{Stair.}\ staircase; \textbf{Dyn.}\ dynamic;
\textbf{Graph} world-graph output; \textbf{View} viewpoint; \textbf{FoR} frame
of reference; \textbf{RTL} reading-direction bias; \textbf{Think.}\ reasoning
effort; \textbf{Halluc.}\ hallucination. \cmark/\xmark\ = present/absent;
\weak/\medium/\strong\ = partial/moderate/strong.
$^{\dag}$~toy/low-diversity simulator. $^{\ddag}$~VLM, not pure-text.
$^{\P}$~Thinking instrumented as direct vs.\ CoT prompting; native long-reasoning
models left to future work. \mentalmap{} combines all eleven dimensions in one
benchmark.
}
\end{table*}

%% file: sections/03_benchmark_design.tex
\section{Benchmark Design}
\label{sec:benchmark_design}

\begin{figure*}[t]
\centering
\includegraphics[width=\linewidth]{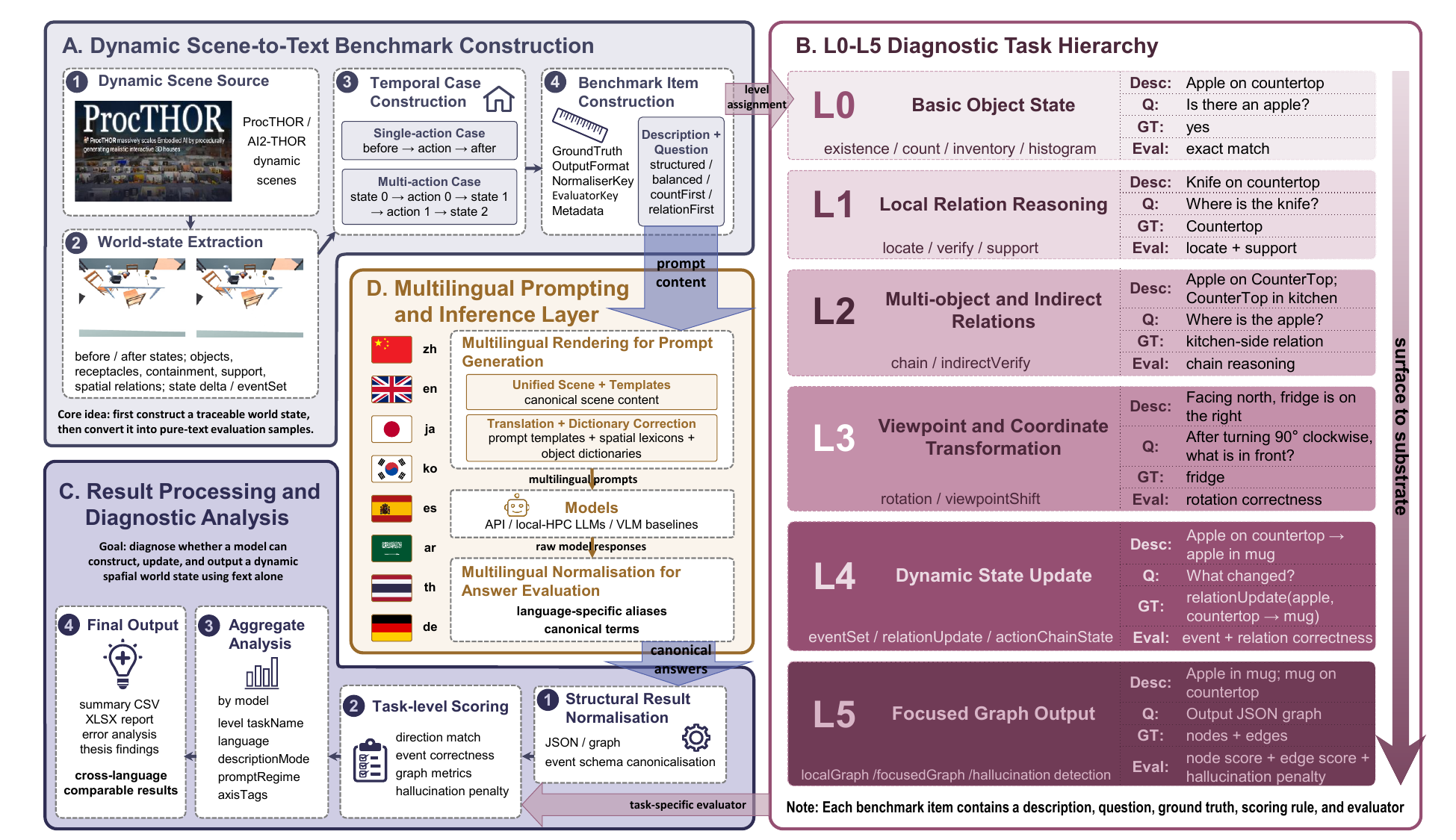}
\caption{
\label{fig:pipeline_staircase}
\textbf{\mentalmap{} overview.}
\textbf{(A)} \emph{Scene-to-text construction.} ProcTHOR scenes are decomposed into traceable world-state artifacts (objects, receptacles, containment, support, spatial relations), then converted into single- and multi-action temporal cases that instantiate benchmark items with canonical ground-truth, output-format, normaliser, and evaluator keys.
\textbf{(B)} \emph{Six-level capability staircase} (L0--L5), from \emph{surface} (atomic object state) to \emph{substrate} (focused graph output); each level lists its task families, an example item, the ground truth, and the scoring rule.
\textbf{(C)} \emph{Result processing.} Structural normalisation of JSON/graph/event outputs, task-level scoring (direction match, event correctness, graph metrics, hallucination penalty), and aggregate analysis along model, level, language, description-mode, prompt-regime, and axis-tag.
\textbf{(D)} \emph{Multilingual prompting and inference.} A unified canonical scene plus prompt templates and spatial-lexicon dictionaries render eight languages (zh, en, ja, ko, es, ar, th, de); language-specific aliases and canonical terms support multilingual answer evaluation.
}
\end{figure*}

\mentalmap{} is designed around three principles: \textbf{(P1)} capability difficulty and diagnostic angle are orthogonal axes that should not be conflated; \textbf{(P2)} pure-text spatial reasoning requires typologically diverse language coverage to expose failure modes; \textbf{(P3)} structured generative output should be scored by partial-credit metrics that distinguish object identification from relation extraction. Figure~\ref{fig:pipeline_staircase} overviews the resulting architecture: a six-stage data pipeline from ProcTHOR scene grounding through canonical extraction and multilingual rendering, instantiating a six-level capability staircase crossed with four orthogonal diagnostic axes. Implementation details, dataset statistics, and prompt templates are in Appendix~\ref{sec:app_design_details}.

\subsection{Two-Tier Capability Structure}
\label{sec:bd_two_tier}

A common pattern in prior spatial-reasoning benchmarks is to add a new ``level'' for each phenomenon, conflating capability difficulty with diagnostic angle. \mentalmap{} enforces the separation through a \textbf{two-tier capability structure}: levels L0--L2 probe \emph{static comprehension}---retrieving previously stated spatial facts---while levels L3--L5 probe \emph{active spatial reasoning}---transforming, simulating, or generating new spatial configurations. The division is cognitively motivated by the recall--construction distinction documented in mental-rotation and egocentric--allocentric studies \citep{shepard1971mental,burgess2006spatial}, and---as we show in Section~\ref{sec:res_staircase}---empirically validated by a universal cliff at the tier boundary.

\noindent\textbf{Tier 1: Static comprehension (L0--L2).} L0 atomic object facts (5 tasks); L1 anchor-local relations (5 tasks); L2 multi-object integration (4 tasks).

\noindent\textbf{Tier 2: Active spatial reasoning (L3--L5).} L3 viewpoint and global direction, structured as three task families: \emph{frame transformation} (\texttt{globalToView}, \texttt{viewToGlobal}), \emph{egocentric manipulation} (\texttt{rotation}, \texttt{viewpointShift}), and a \emph{consistency probe} (\texttt{rotationCheck}). L4 dynamic state update (single-event, multi-event, multi-step chain; 13 tasks total). L5 generative world-graph output, with four sub-tasks (static, dynamic, counterfactual, local slice). Full task definitions in Appendix~\ref{sec:app_design_details}.

\subsection{Four Diagnostic Axes}
\label{sec:bd_axes}

\input{tables/taxonomy_matrix}

Orthogonal to the staircase, \mentalmap{} instruments four diagnostic axes (Table~\ref{tab:taxonomy_matrix}). \textbf{Frame of reference (FoR):} each spatial relation is rendered in three variants---relative, absolute, intrinsic \citep{levinson2003space,tenbrink2011reference}---enabling within-model frame-shift analysis (Appendix Table~\ref{tab:app_for_examples} for per-language templates). \textbf{Reading-direction bias (RTL):} two mention orders test anchor-preference patterns documented for human reasoners \citep{vanderhenst2008influence}. \textbf{Reasoning effort:} each task is presented under paired direct and reasoning (chain-of-thought) prompts, yielding per-cell reasoning-prompt deltas that reveal latent vs.\ readily-available competence (Section~\ref{sec:res_reasoning_prompt}). \textbf{Hallucination:} L5 graph generation is scored by decomposed sub-metrics---hallucinated nodes, missing nodes, hallucinated edges, relation flips---exposing the node--edge dissociation that aggregate F1 obscures. The axes are not applied uniformly: FoR is degenerate at L0; hallucination is most diagnostically rich at L5. Table~\ref{tab:taxonomy_matrix} formalizes the applicability matrix.

\subsection{Construction and Metrics}
\label{sec:bd_construction}

\noindent\textbf{Scene grounding.} 100 ProcTHOR \citep{deitke2022procthor} houses (median 5--7 rooms) provide ground-truth scene graphs; AI2-THOR \citep{kolve2017ai2thor} provides 17 high-level physical actions. Each scene--action episode is decomposed into three language-independent canonical artifacts (world state, trajectory, transition log) that instantiate tasks across all six levels.

\noindent\textbf{Language coverage and scale.} Eight typologically diverse languages span all three Levinsonian frame classes and combine LTR/RTL writing systems: en (relative), zh/ja/ko (intrinsic), es/de (relative), ar (absolute), th (mixed). A structured-text control isolates spatial reasoning from language quality. The benchmark instantiates 39 task families across the six levels, rendered in nine conditions (eight natural languages plus the structured-text control) under three description modes and two prompt regimes, yielding 1{,}950 evaluation cells and 47{,}550 scored items per model. All non-English renderings were native-speaker validated, with validators instructed to rewrite (not only approve) translations that read unnaturally; the structured-text control provides an additional check against translation-quality effects. Construction pipeline and per-level case breakdowns are in Appendix~\ref{sec:app_design_details} (Table~\ref{tab:app_dataset_counts}).

\noindent\textbf{Composite scoring.} Two complementary composites per cell: a \emph{strict composite} (unweighted macro-average of task-level pass rates) for L0--L4, and a \emph{partial-credit composite} (case-weighted graph F1) for L5. Section~\ref{sec:res_metric} shows the two composites diverge substantially on several models---reporting both protects against metric-choice artifacts. Task-level metrics including L4 event precision/recall (matching against the gold action--state transition log), L5c counterfactual delta F1 (graph F1 restricted to changed nodes/edges), and L5 node/edge F1 (set F1 over predicted nodes and containment edges) are defined in Appendix~\ref{sec:app_metrics}.

%% file: tables/taxonomy_matrix.tex
\begin{table}[t]
\centering
\scriptsize
\setlength{\tabcolsep}{3pt}
\renewcommand{\arraystretch}{1.1}

\begin{tabular}{l *{4}{>{\centering\arraybackslash}p{0.85cm}}}
\toprule
\textbf{Level} & \textbf{FoR} & \textbf{RTL} & \textbf{Think.} & \textbf{Halluc.} \\
\midrule

\rowcolor{rowlavender}
\multicolumn{5}{l}{\textbf{\textit{(A) Static comprehension (L0--L3)}}} \\

L0 Atomic            & \dash    & \weak    & \weak    & \weak    \\
L1 Anchor            & \weak    & \weak    & \weak    & \weak    \\
L2 Multi-Object      & \weak    & \weak    & \weak    & \weak    \\
L3 Viewpoint         & \strong  & \weak    & \weak    & \dash    \\

\rowcolor{rowpink}
\multicolumn{5}{l}{\textbf{\textit{(B) Dynamic state update (L4)}}} \\

L4a Single-Event     & \dash    & \dash    & \weak    & \medium  \\
L4b Multi-Event      & \dash    & \dash    & \weak    & \medium  \\
L4c Multi-Step Chain & \dash    & \dash    & \medium  & \weak    \\

\rowcolor{rowbeige}
\multicolumn{5}{l}{\textbf{\textit{(C) Generative world-graph output (L5)}}} \\

L5a Static Graph     & \dash    & \dash    & \weak    & \strong  \\
L5b Dynamic Graph    & \dash    & \dash    & \weak    & \strong  \\
L5c Counterfactual   & \dash    & \dash    & \weak    & \strong  \\
L5d Local Slice      & \dash    & \dash    & \weak    & \strong  \\

\bottomrule
\end{tabular}
\caption{
\label{tab:taxonomy_matrix}
\textbf{Applicability matrix for \mentalmap{}'s two-tier architecture.}
Rows: capability staircase grouped into static comprehension (L0--L3),
dynamic state (L4), and generative graph (L5). Columns: \textbf{FoR},
\textbf{RTL}, \textbf{Think.}, \textbf{Halluc.}\ diagnostic axes.
\dash\ N/A; \weak\ supporting; \medium\ primary; \strong\ core.
}
\end{table}

%% file: sections/04_experiments_and_results.tex
\section{Experiments and Results}
\label{sec:experiments_results}

\noindent\textbf{Models and protocol.} We evaluate thirteen LLMs: four frontier closed-source (GPT-4o~\citep{openai2024gpt4o}, Gemini-2.5-Flash~\citep{google2025gemini25}, DeepSeek-V4-Flash~\citep{deepseek2025r1}, Qwen3.5-Flash~\citep{qwen35flash2025}); five open-weight 7--10B (Llama-3.1-8B~\citep{grattafiori2024llama3}, Qwen2.5-7B~\citep{qwen2024qwen25}, Mistral-7B-v0.3~\citep{jiang2023mistral7b}, Gemma-2-9B~\citep{gemma2-2024}, Falcon3-10B~\citep{falcon3-2025}); Qwen2.5-32B and Gemma-3-27B-it as scale controls for F5; SmolLM2-1.7B~\citep{allal2025smollm2} as a scale lower bound establishing the sub-cliff regime; and Qwen2.5-VL-3B~\citep{qwen25vl-2025} as a vision-language text-only ablation. Open models are served via vLLM~\citep{vllm2023} on A100 80GB with deterministic decoding. Each task--language cell is evaluated under direct and chain-of-thought prompts, yielding the per-cell reasoning-prompt delta (\S\ref{sec:res_reasoning_prompt}); diagnostic axes use specialized protocols (Appendix~\ref{sec:app_eval_protocol}). The cross-axis summary (Table~\ref{tab:cross_axis_summary}) consolidates evidence for F2--F4 in a single half-page reference.

\subsection{Capability Staircase and the Universal L3 Cliff}
\label{sec:res_staircase}

\input{tables/main_results}

\begin{figure}[t]
\centering
\includegraphics[width=\linewidth]{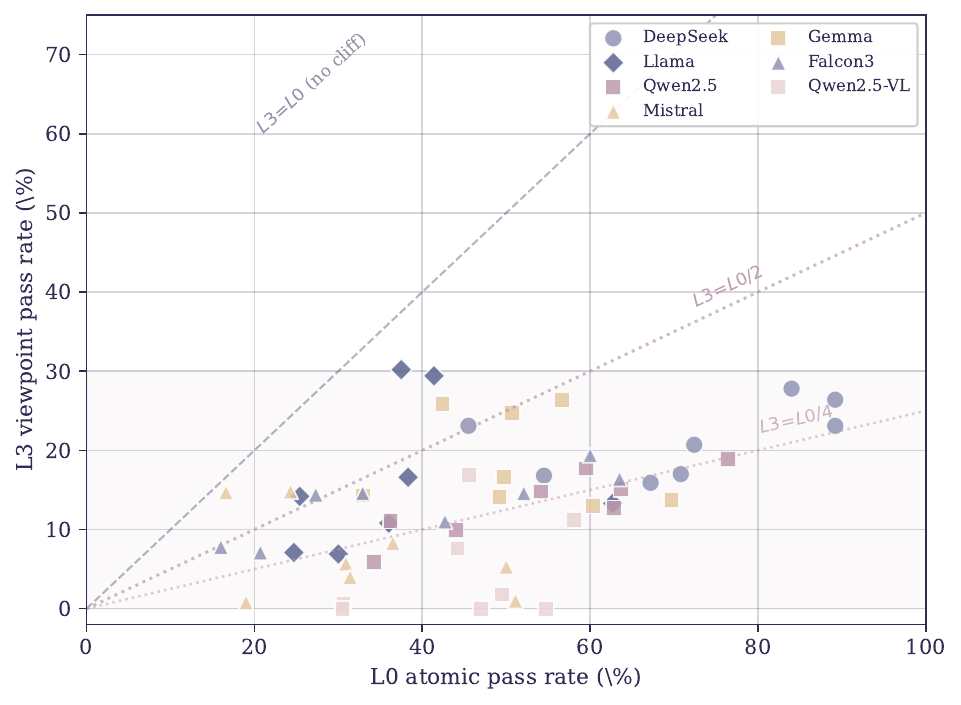}
\caption{
\label{fig:l0_l3_cliff}
\textbf{Universal L3 viewpoint cliff (F1).} Each point is one (model, language) pair. All points lie below $L3\!=\!L0/2$; most below $L3\!=\!L0/4$.
}
\end{figure}

Table~\ref{tab:main_results} reports composite scores per (model, language, level). The staircase is non-monotonic: all twelve cliff-diagnostic models (i.e., excluding the SmolLM2-1.7B sub-cliff lower bound) show a sharp drop at L3 viewpoint and partial recovery at L4--L5. Figure~\ref{fig:l0_l3_cliff} visualizes this directly: every (model, language) pair in the cliff-diagnostic regime lies below the $L3\!=\!L0/2$ reference line, with the majority below $L3\!=\!L0/4$. The cliff is not a gradient: DeepSeek-V4-Flash retains $\sim$89\% at L0 atomic facts but collapses to $\sim$23\% at L3 viewpoint in English---a four-fold drop within a single tier transition. No model in any language shows a smooth degradation across the static-to-active boundary; the staircase \emph{breaks} at L3, universally and discretely. Static-comprehension levels L0--L2 and the full per-model breakdown across all six levels are reported in Appendix Tables~\ref{tab:app_l0_l2} and~\ref{tab:app_per_model_results}.

\input{tables/cross_axis_summary}

\subsection{The L3 Cliff Is Sub-Task Structured (F2)}
\label{sec:res_l3_cliff}

The left block of Table~\ref{tab:cross_axis_summary} decomposes L3 along the three task families. The \emph{consistency-probe} family (\texttt{rotationCheck}) admits boolean guessing: Qwen2.5-7B reaches 90.0\% in English but 0--3\% in most non-English variants---language-specific pattern match rather than transferred capability. The \emph{egocentric-manipulation} family is partially solvable by open-weight models (Llama ja 48.9\%, Gemma en 36.7\%) but near-zero for several frontier closed-source models. The \emph{frame-transformation} family is most consistently solvable by frontier closed-source models (GPT-4o 21--52\%) while most open-weight models score 0\% in seven of eight languages. Aggregate L3 thus mixes genuine viewpoint-transformation failure with metric ceiling effects on the boolean probe; family-stratified reporting (Appendix Table~\ref{tab:app_l3_subtask_lang}) enables finer-grained diagnosis. Recomputing L3 without \texttt{rotationCheck} drops scores by an average of 3.0pp but $L3\!<\!L0/2$ still holds in 69 of 72 cells, confirming F1 is not a boolean-probe artefact.

\subsection{Chain-of-Thought Is Not a Universal Booster (F3)}
\label{sec:res_reasoning_prompt}

\begin{figure}[t]
\centering
\includegraphics[width=\linewidth]{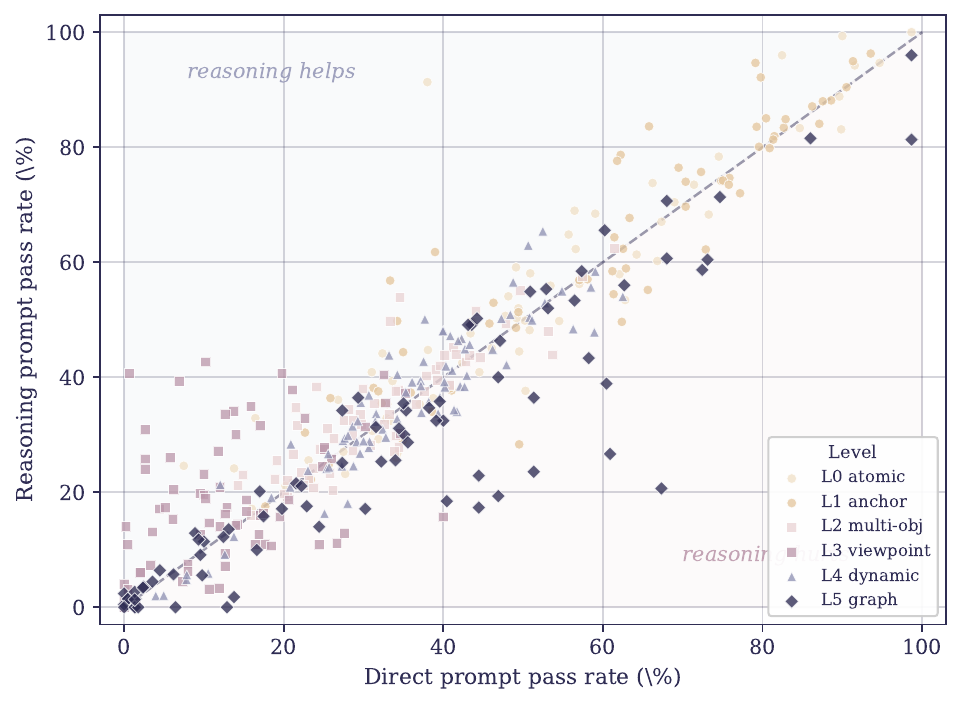}
\caption{
\label{fig:direct_reasoning}
\textbf{Reasoning-prompt effect is level-stratified (F3).} Each point is one (model, language, level) cell. Diagonal: no effect. L3 cells cluster above; high-direct L5 cells cluster on or below the diagonal.
}
\end{figure}

The middle block of Table~\ref{tab:cross_axis_summary} shows that the reasoning-prompt effect is both model- and level-dependent. DeepSeek-V4-Flash gains the most: $+32.4$pp at L3 (10.2 $\to$ 42.7\%), with smaller positive gains across L0--L4. Llama-3.1-8B also benefits at L0--L3 ($+$12 to $+$18pp). Conversely, GPT-4o, Gemini, Qwen2.5-7B, and Gemma show \emph{negative} deltas at L3--L5, with Qwen2.5-7B losing 15.6pp at L3 and most open-weight models losing 14--27pp at L5. The L5 degradation has a concrete mechanism: JSON validity under reasoning collapses for 11 of 13 evaluated models (Mistral-7B from 70.4\% to 7.3\%, the most extreme; Appendix Table~\ref{tab:app_json_validity}); only DeepSeek-V4 ($+5.2$pp) and the sub-cliff SmolLM2-1.7B ($+2.7$pp) show small positive JSON-validity deltas under reasoning, and DeepSeek is the same model that shows the largest positive L3 delta. DeepSeek's L3 gain plausibly reflects multi-step decoding allowing intermediate reference resolution, while the open-weight L5 collapse reflects reasoning preambles competing with strict JSON token budgets---each delta is a property of the (model, prompt-template) pair rather than an intrinsic capacity claim.

\subsection{Node--Edge Dissociation and Strict-vs-Partial Metrics (F4)}
\label{sec:res_metric}

\input{tables/l4_counterfactual}

The right block of Table~\ref{tab:cross_axis_summary} reports L5 node F1 and edge F1; the two dissociate strongly in \emph{every} evaluated model, with all 13 models showing node F1 strictly greater than edge F1 (mean gaps $+27.5$pp for 7--10B open-weight, $+18.6$pp for the 27--32B controls, $+15.8$pp for closed-source, $+16.2$pp for the sub-cliff SmolLM2-1.7B; max $+33.6$pp on Qwen2.5-7B; full per-(model, sub-task, language) scatter in Figure~\ref{fig:app_node_edge}). The dissociation is most pronounced on \texttt{staticFocusedGraph} (Appendix Table~\ref{tab:app_l5_graph_diagnostics}). The same strict-vs-partial split applies to L4: Falcon3-10B reaches 53.8\% strict pass but only 5.6\% event precision (template-matching, not event tracking), and on L5c counterfactual most open-weight models recover 80--88\% graph F1 despite only 50--60\% strict pass (Table~\ref{tab:l4_counterfactual}). Reporting both metrics protects against artifacts that single-composite benchmarks miss.

\noindent\textbf{Scale does not rescue the cliff; L5 convergence reflects fine-tuning (F5).} Across L0--L4 closed-source leads 7--10B open-weight by 5--26~pp on average (Table~\ref{tab:main_results}); at L5 the two groups score within $\sim$1pp. Both mid-scale open-weight controls---Qwen2.5-32B (Overall L3$\!=\!13.7$) and Gemma-3-27B-it (Overall L3$\!=\!22.4$)---exhibit the L3 cliff with $L3/L0\!<\!0.5$ in all 8 languages, while reaching L5 Overall $71.6$ and $84.0$ respectively, with Gemma-3-27B's L5 matching closed-source ceilings and Qwen2.5-32B's L5c counterfactual matching GPT-4o (97.8/98.2; Table~\ref{tab:l4_counterfactual}). At the lower bound, SmolLM2-1.7B sits below the cliff-diagnostic regime: L0 collapses to $\sim$29\% (near random for several languages) and L3 cannot be cleanly separated from L0 (the model's only non-zero L3 rates in ar/th match the 20\% random baseline of \texttt{rotationCheck}). The cliff is therefore a property of the mid-capability regime ($\geq$7B); it persists across a $\sim$19$\times$ parameter range (1.7B to 32B) and at no point does parameter count save L3---consistent with fine-tuning regime over parameter count.

\subsection{Multilingual Fingerprints Recover Script Typology (F6)}
\label{sec:res_cross_lang}

\begin{figure}[!t]
\centering
\includegraphics[width=\linewidth]{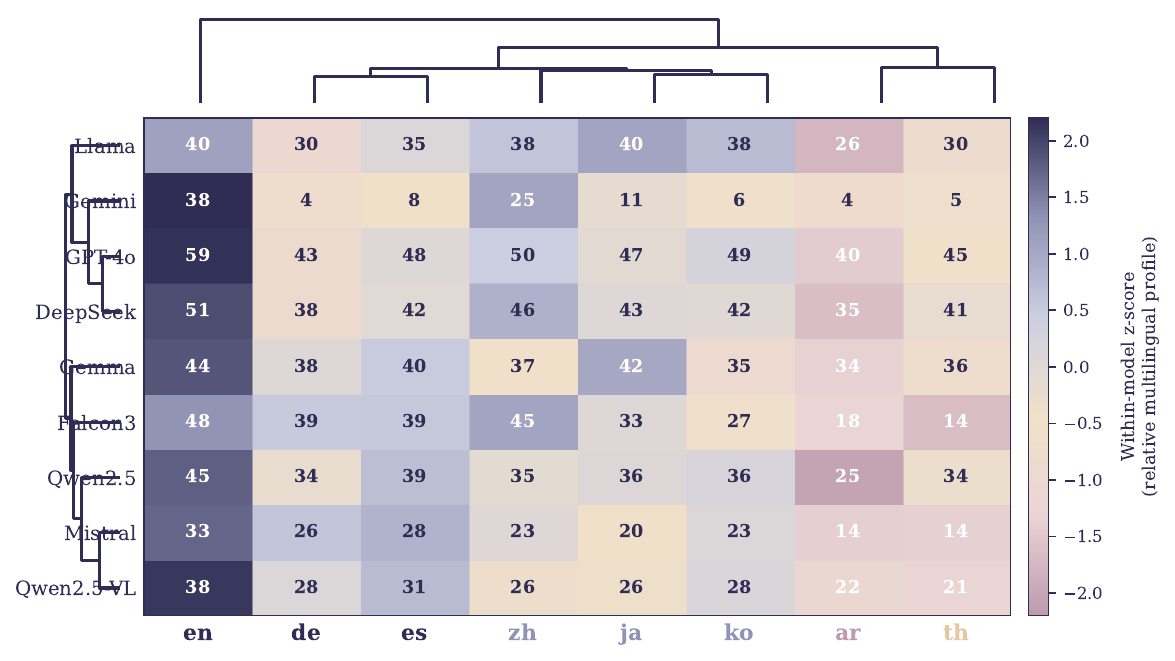}
\caption{
\label{fig:multilingual_fingerprint}
\textbf{Multilingual fingerprints (F6).} Heatmap of the nine principal LLMs $\times$ eight languages (Qwen2.5-32B scale control and Qwen2.5-VL ablation excluded from this view); cells encode within-model z-score (color) and absolute pass rate (number, mean of L3--L5). The two \emph{dendrograms} are hierarchical-clustering trees: closer branches share more similar performance profiles. The top tree clusters languages (en/de/es vs.\ zh/ja/ko vs.\ ar/th), reproducing the Latin/CJK/Arabic--Thai script typology without external features; the left tree clusters models into a narrow-profile regime (Llama, Gemini) and a broad-profile regime (the remaining seven).
}
\end{figure}

Per-language averages collapse model-specific multilingual profiles into a single number. To surface the underlying structure, we cluster the (model, language) matrix on both axes via average-linkage on within-model z-score profiles (Figure~\ref{fig:multilingual_fingerprint}; cell values overlay absolute pass rates).

\input{tables/script_category}

\noindent\textbf{Language dendrogram reproduces script typology; model fingerprints reveal two regimes.} The top dendrogram---built purely from z-score profiles with no external linguistic features---reproduces the Latin / CJK / (Arabic, Thai) typological grouping; the Arabic--Thai joint cluster suggests the cross-language gap is consistent with \emph{training-data script coverage} rather than typology. The left dendrogram separates models into narrow-profile (Llama-3.1-8B, Gemini-2.5-Flash) and broad-profile (the remaining seven) groups; GPT-4o clusters with the 7--10B open-weight models while Llama clusters with Gemini---multilingual coverage is a fine-tuning property, not a scale property. Table~\ref{tab:script_category} aggregates along the groups, and PCA and anchor-preference views (Figure~\ref{fig:app_pca}, Appendix Table~\ref{tab:app_anchor_preference}) corroborate the script-coverage gradient.

\subsection{Cross-Lingual Human Pilot Reproduces the Cliff (F7)}
\label{sec:res_human_pilot}

F1--F6 describe \emph{what} models fail at but leave open \emph{why}. A capability account predicts fluent humans should not collapse on the same items; a modality account predicts they should. We ran a cross-lingual human pilot under the identical pure-text protocol: native speakers across all eight evaluation languages (N$\geq$3 per language) attempted a stratified random sample from each language split through a terminal harness identical to the LLM prompt, no scratchpad, strict skip$=$0 scoring; items carry an a~priori \textsc{Easy}/\textsc{Medium}/\textsc{Hard} tier from referent count and chain depth (Appendix~\ref{sec:app_human_pilot}).

\noindent The result is consistent with the modality account, and the cross-lingual replication is strong. \textbf{L3 collapses to 0\% in all eight languages}; \textbf{L4 is language-invariant} (range 35.4--45.2\%, mean 41.1, std 3.3pp). All skipped items are L3 frame-transformations or multi-event L4 tasks, precisely where models also collapse (Table~\ref{tab:main_results}). The language-invariance rules out language-specific accounts: the cliff is a property of \emph{pure-text presentation}, with the working-memory bottleneck discussed in Section~\ref{sec:conclusion} as the most parsimonious mechanism.

%% file: tables/main_results.tex
%

\providecommand{\sep}{\textcolor{black!35}{/}}
\providecommand{\nodata}{\textcolor{black!45}{\textsc{n/a}}}

\begin{table*}[t]
\centering
\scriptsize
\setlength{\tabcolsep}{1pt}
\renewcommand{\arraystretch}{1.25}
\resizebox{\textwidth}{!}{%
\begin{tabular}{l cccccccc c c}
\toprule
& \multicolumn{8}{c}{\textbf{Language}}
& \multicolumn{2}{c}{} \\
\cmidrule(lr){2-9}
\textbf{Model}
& \textbf{en} & \textbf{zh} & \textbf{ja} & \textbf{ko} & \textbf{es} & \textbf{ar} & \textbf{th} & \textbf{de}
& \textbf{Struct.} & \textbf{Overall} \\
\midrule
\rowcolor{rowlavender}
\multicolumn{11}{l}{\textbf{\textit{(A) Frontier closed-source}}} \\
GPT-4o~\citep{openai2024gpt4o}
  & 32.4\sep52.4\sep93.3
  & 30.9\sep44.3\sep75.8
  & 24.2\sep40.5\sep77.6
  & 36.3\sep40.8\sep70.7
  & 23.3\sep39.2\sep81.9
  & 36.0\sep29.8\sep55.1
  & 26.3\sep37.8\sep71.6
  & 24.8\sep35.9\sep69.1
  & 34.2\sep61.5\sep99.4
  & 29.3\sep40.1\sep74.4 \\
Gemini-2.5-Flash~\citep{google2025gemini25}
  & 21.3\sep61.2\sep31.6
  & 17.5\sep47.9\sep9.2
  & 15.1\sep7.6\sep9.7
  & 1.3\sep10.2\sep8.0
  & 4.2\sep13.0\sep10.0
  & 2.4\sep3.6\sep6.9
  & 2.6\sep8.4\sep7.5
  & 5.2\sep4.0\sep8.0
  & 29.3\sep49.3\sep38.7
  & 8.7\sep19.5\sep11.4 \\
DeepSeek-V4-Flash~\citep{deepseek2025r1}
  & 26.4\sep56.9\sep71.0
  & 27.8\sep45.4\sep66.2
  & 23.1\sep40.5\sep64.3
  & 20.7\sep40.3\sep65.7
  & 15.9\sep45.0\sep65.4
  & 23.1\sep29.8\sep51.1
  & 17.0\sep45.4\sep62.0
  & 16.8\sep41.5\sep56.4
  & 22.0\sep66.3\sep72.6
  & 21.4\sep43.1\sep62.8 \\
Qwen3.5-Flash~\citep{qwen35flash2025}
  & 14.2\sep65.2\sep75.6
  & 14.1\sep50.2\sep77.5
  & 16.5\sep41.0\sep71.2
  & 9.4\sep37.3\sep64.4
  & 7.8\sep42.0\sep71.6
  & 14.0\sep29.8\sep52.6
  & 12.3\sep40.8\sep74.4
  & 5.6\sep42.9\sep70.1
  & 14.1\sep74.2\sep77.2
  & 11.7\sep43.7\sep69.7 \\
\midrule
\rowcolor{rowbeige}
\multicolumn{11}{l}{\textbf{\textit{(B$^{\flat}$) Sub-cliff regime ($\leq$2B)}}} \\
SmolLM2-1.7B-Instruct~\citep{allal2025smollm2}
  & 0.0\sep37.6\sep41.4
  & 4.3\sep27.7\sep4.5
  & 0.0\sep26.2\sep14.4
  & 0.0\sep26.4\sep5.7
  & 0.0\sep25.4\sep22.4
  & 20.0\sep20.5\sep8.7
  & 20.0\sep18.1\sep3.0
  & 0.0\sep23.8\sep20.3
  & 0.9\sep43.1\sep26.2
  & 5.5\sep25.7\sep15.0 \\
\midrule
\rowcolor{rowbeige}
\multicolumn{11}{l}{\textbf{\textit{(B) Open-weight (7--10B)}}} \\
Llama-3.1-8B-Instruct~\citep{grattafiori2024llama3}
  & 13.3\sep45.1\sep62.0
  & 16.6\sep35.2\sep60.8
  & 30.2\sep30.5\sep59.0
  & 29.4\sep26.7\sep58.5
  & 10.8\sep30.1\sep64.2
  & 14.2\sep16.7\sep46.2
  & 6.9\sep32.6\sep51.5
  & 7.1\sep33.4\sep48.2
  & 13.3\sep52.0\sep76.6
  & 16.1\sep31.3\sep56.3 \\
Qwen2.5-7B-Instruct~\citep{qwen2024qwen25}
  & 18.9\sep41.0\sep73.8
  & 12.7\sep35.5\sep56.3
  & 15.1\sep27.1\sep65.2
  & 14.8\sep28.0\sep66.7
  & 17.7\sep35.3\sep64.0
  & 5.9\sep19.9\sep48.6
  & 9.9\sep26.1\sep65.6
  & 11.1\sep33.5\sep58.1
  & 16.0\sep55.8\sep83.8
  & 13.3\sep30.8\sep62.3 \\
Mistral-7B-Instruct-v0.3~\citep{jiang2023mistral7b}
  & 1.0\sep53.5\sep44.2
  & 5.7\sep35.3\sep26.8
  & 5.3\sep28.3\sep25.1
  & 8.3\sep30.9\sep29.8
  & 14.8\sep31.7\sep36.3
  & 0.8\sep22.6\sep19.2
  & 4.0\sep20.3\sep18.5
  & 14.7\sep31.4\sep32.3
  & 2.0\sep53.4\sep57.2
  & 6.8\sep31.8\sep29.0 \\
Gemma-2-9B-it~\citep{gemma2-2024}
  & 13.7\sep48.9\sep70.1
  & 13.0\sep39.4\sep58.4
  & 26.4\sep39.3\sep58.9
  & 16.6\sep33.8\sep55.6
  & 24.7\sep39.4\sep55.6
  & 14.3\sep26.2\sep61.2
  & 14.1\sep32.7\sep61.2
  & 25.9\sep36.2\sep53.2
  & 19.7\sep65.7\sep90.4
  & 18.6\sep37.0\sep59.3 \\
Falcon3-10B-Instruct~\citep{falcon3-2025}
  & 16.4\sep44.5\sep82.0
  & 19.4\sep36.8\sep78.4
  & 14.6\sep25.0\sep58.9
  & 11.0\sep21.2\sep48.3
  & 14.6\sep32.7\sep69.7
  & 7.8\sep18.3\sep29.2
  & 7.1\sep12.7\sep21.3
  & 14.4\sep31.4\sep70.4
  & 4.0\sep56.4\sep82.8
  & 13.2\sep27.8\sep57.3 \\
\midrule
\rowcolor{rowbeige}
\multicolumn{11}{l}{\textbf{\textit{(B$'$) Mid-scale open-weight (27--32B)}}} \\
Qwen2.5-32B-Instruct~\citep{qwen2024qwen25}
  & 14.7\sep56.2\sep95.7
  & 23.3\sep47.3\sep80.7
  & 8.9\sep36.9\sep61.3
  & 6.6\sep37.6\sep61.2
  & 17.7\sep40.7\sep68.5
  & 10.2\sep25.3\sep59.8
  & 18.0\sep40.9\sep72.3
  & 9.9\sep39.6\sep73.4
  & 9.7\sep68.7\sep89.3
  & 13.7\sep40.6\sep71.6 \\
Gemma-3-27B-it~\citep{gemma3-2025}
  & 28.7\sep70.6\sep87.5
  & 17.3\sep68.7\sep77.6
  & 28.7\sep60.8\sep81.1
  & 35.3\sep57.8\sep81.3
  & 18.0\sep65.3\sep97.2
  & 26.0\sep49.3\sep75.4
  & 10.7\sep57.7\sep93.8
  & 14.7\sep59.8\sep78.4
  & \nodata
  & 22.4\sep61.3\sep84.0 \\
\midrule
\rowcolor{rowbeige}
\multicolumn{11}{l}{\textbf{\textit{(B$''$) Vision-language (text-only ablation)}}} \\
Qwen2.5-VL-3B-Instruct~\citep{qwen25vl-2025}
  & 11.2\sep37.6\sep66.4
  & 7.7\sep28.1\sep50.1
  & 1.8\sep26.1\sep49.6
  & 16.9\sep25.9\sep42.6
  & 7.6\sep31.2\sep55.5
  & 0.6\sep26.7\sep39.0
  & 1.4\sep20.9\sep43.4
  & 4.1\sep29.5\sep54.8
  & 10.7\sep45.0\sep58.0
  & 6.4\sep28.2\sep50.2 \\
\midrule
\rowcolor{rowpink}
\multicolumn{11}{l}{\textbf{\textit{(C) Human pilot (pure-text, no scratchpad)}}} \\
Human (N=3+ per lang)
  & 0.0\sep42.3\sep\nodata
  & 0.0\sep41.7\sep\nodata
  & 0.0\sep45.2\sep\nodata
  & 0.0\sep44.6\sep\nodata
  & 0.0\sep38.6\sep\nodata
  & 0.0\sep37.6\sep\nodata
  & 0.0\sep35.4\sep\nodata
  & 0.0\sep43.5\sep\nodata
  & \nodata
  & 0.0\sep41.1\sep\nodata \\
\bottomrule
\end{tabular}%
}
\caption{
\label{tab:main_results}
\textbf{Main results: capability staircase across eight languages.}
Each cell shows three composites \texttt{L3\,/\,L4\,/\,L5} (\%) where L3 is
viewpoint, L4 dynamic state, L5 generative world-graph. \textbf{Struct.}\ is
the structured-text control. \textbf{Overall} is the unweighted cross-language
mean over the eight natural languages. L3 and L4 are macro-averages of
\texttt{passRate} (strict binary correctness after canonical normalisation,
$n\!\ge\!10$ per task); L5 is the macro-average of \texttt{averageScore},
the L5-composite partial-credit metric defined in Appendix~\ref{sec:app_metrics}
(weighted combination of JSON/schema validity, node F1, edge F1, and relation
correctness; coefficients $0.05/0.05/0.20/0.55/0.15$).
\textsc{n/a} marks cells with insufficient data.
Row groups are colour-coded: \textbf{(A)} closed-source frontier (lavender),
\textbf{(B$^{\flat}$)} sub-cliff regime ($\leq$2B) where L0 is near random,
included as a scale lower bound (beige), \textbf{(B)} open-weight 7--10B
(beige), \textbf{(B$'$)} mid-scale open-weight 27--32B serving as scale
controls for F5 (beige), \textbf{(B$''$)} vision-language model evaluated
text-only as ablation (beige), \textbf{(C)} human pilot (pink).
For SmolLM2-1.7B, the L3 rates of 20\% in ar/th correspond to the random
baseline of the \texttt{rotationCheck} boolean sub-task; the model sits
below the cliff-diagnostic regime, where recall (L0) and construction (L3)
cannot be cleanly separated.
The human row reports native speakers across all eight evaluation languages
(N$\geq$3 participants per language) on a stratified random sample drawn from
the matched-difficulty L3 (viewpoint) and L4 (dynamic state) cells; pure-text
presentation with no scratchpad or external memory aid; strict scoring with
skip$=$0. The L4 cell pools event/undo/counterfactual sub-tasks; L5 and the
structured-text control are out of scope. The L4 result is remarkably
language-invariant (range 35.4--45.2\%, std=3.3pp), and L3 collapses to 0\%
across all eight languages---direct evidence that the cliff is a property
of pure-text presentation rather than any specific language.
}
\end{table*}

%% file: tables/cross_axis_summary.tex
\begin{table}[t]
\centering
\scriptsize
\setlength{\tabcolsep}{2.5pt}
\renewcommand{\arraystretch}{1.15}
\resizebox{\columnwidth}{!}{%
\begin{tabular}{l ccc | cccccc | cc}
\toprule
& \multicolumn{3}{c|}{\textbf{L3 sub-task \%}}
& \multicolumn{6}{c|}{\textbf{Reasoning $\Delta$ per level (pp)}}
& \multicolumn{2}{c}{\textbf{L5 graph}} \\
\textbf{Model}
& \textbf{rot} & \textbf{rChk} & \textbf{v$\to$g}
& \textbf{L0} & \textbf{L1} & \textbf{L2} & \textbf{L3} & \textbf{L4} & \textbf{L5}
& \textbf{nF1} & \textbf{eF1} \\
\midrule
\rowcolor{rowlavender}
\multicolumn{12}{l}{\textbf{\textit{(A) Frontier closed-source}}} \\
GPT-4o~\citep{openai2024gpt4o}            & 1.9  & 76.9 & 45.4 &  --  &  --  &  --  & $-$4.4  & $-$7.9 & $-$4.4  & 96.8 & 88.2 \\
Gemini-2.5-Flash~\citep{google2025gemini25}  & 8.3  & 4.2  & 56.2 &  --  &  --  &  --  & $-$14.8 & $-$8.5 & $-$0.5  & 33.6 & 24.9 \\
DeepSeek-V4-Flash~\citep{deepseek2025r1} & 0.0  & 50.0 & 1.1  & +13.6 & +4.6 & +5.4 & +32.4 & +12.9 & +5.3 & 53.5 & 44.4 \\
Qwen3.5-Flash~\citep{qwen35flash2025}    & 0.0  & 0.0  & 31.2 & +13.2 & $-$2.4 & $-$3.8 & +23.8 & +0.0 & $-$19.6 & 98.7 & 98.4 \\
\midrule
\rowcolor{rowbeige}
\multicolumn{12}{l}{\textbf{\textit{(B$^{\flat}$) Sub-cliff regime ($\leq$2B)}}} \\
SmolLM2-1.7B~\citep{allal2025smollm2}     & 0.0  & 0.0  & 0.0  & +20.7 & +0.2 & $-$0.4 & +2.3 & +6.3 & $-$6.4   & 45.8 & 29.6 \\
\midrule
\rowcolor{rowbeige}
\multicolumn{12}{l}{\textbf{\textit{(B) Open-weight (7--10B)}}} \\
Llama-3.1-8B~\citep{grattafiori2024llama3}       & 2.2  & 0.0  & 28.9 & +12.5 & +17.9 & +16.4 & +14.2 & $-$2.6 & $-$7.6  & 87.4 & 53.1 \\
Qwen2.5-7B~\citep{qwen2024qwen25}         & 25.6 & 90.0 & 13.3 & +3.9 & $-$5.2 & +1.4 & $-$15.6 & $-$1.3 & +1.1   & 88.1 & 66.2 \\
Mistral-7B-v0.3~\citep{jiang2023mistral7b}    & 0.0  & 7.8  & 0.0  &  --  &  --  &  --  &  --  &  --  &  --                 & 75.7 & 43.2 \\
Gemma-2-9B~\citep{gemma2-2024}         & 36.7 & 7.8  & 23.3 & +1.5 & $-$1.0 & +0.9 & $-$5.6 & +0.2 & $-$21.6     & 89.7 & 70.4 \\
Falcon3-10B~\citep{falcon3-2025}        & 0.0  & 18.9 & 41.1 & $-$6.6 & $-$2.3 & +3.6 & +8.9 & +2.9 & $-$13.8   & 76.7 & 66.1 \\
\midrule
\rowcolor{rowbeige}
\multicolumn{12}{l}{\textbf{\textit{(B$'$) Mid-scale open-weight (27--32B)}}} \\
Qwen2.5-32B~\citep{qwen2024qwen25}       & 7.8  & 10.0 & 26.7 & +5.0 & +1.6 & $-$1.8 & +3.6 & $-$10.8 & +0.4   & 94.6 & 92.0 \\
Gemma-3-27B-it~\citep{gemma3-2025}       & 36.7 & 66.7 & 3.3  & +8.9 & $-$0.2 & +18.0 & +4.0 & +5.1 & +4.2   & 99.1 & 78.4 \\
\midrule
\rowcolor{rowbeige}
\multicolumn{12}{l}{\textbf{\textit{(B$''$) Vision-language (text-only ablation)}}} \\
Qwen2.5-VL-3B~\citep{qwen25vl-2025}      & 0.0  & 13.1 & 12.6 & $-$9.3 & $-$0.7 & +0.5 & +12.1 & +3.1 & $-$0.8    & 54.5 & 30.3 \\
\bottomrule
\end{tabular}%
}
\caption{
\label{tab:cross_axis_summary}
\textbf{Cross-axis findings (English, direct prompt where applicable).}
\emph{Left}: L3 sub-task pass rates---\textbf{rot} (rotation), \textbf{rChk}
(rotationCheck), \textbf{v$\to$g} (viewToGlobal). \emph{Middle}: reasoning $-$
direct prompt delta per level. \emph{Right}: L5 node F1 / edge F1 averaged over
four L5 sub-tasks.
}
\end{table}

%% file: tables/l4_counterfactual.tex

\begin{table}[t]
\centering
\scriptsize
\setlength{\tabcolsep}{4pt}
\renewcommand{\arraystretch}{1.2}
\begin{tabular}{l ccc | cc}
\toprule
& \multicolumn{3}{c|}{\textbf{L4 dynamic state (en, direct)}}
& \multicolumn{2}{c}{\textbf{L5c counterfactual}} \\
\textbf{Model}
& \textbf{pass\%} & \textbf{evP\%} & \textbf{evR\%}
& \textbf{pass\%} & \textbf{graphF1\%} \\
\midrule
\rowcolor{rowlavender}
\multicolumn{6}{l}{\textbf{\textit{(A) Frontier closed-source}}} \\
GPT-4o~\citep{openai2024gpt4o}            & 67.0 & 61.7 & 46.7 & 74.6 & 96.2 \\
Gemini-2.5-Flash~\citep{google2025gemini25}  & 68.8 & 44.3 & 31.0 & 83.3 &  --  \\
DeepSeek-V4-Flash~\citep{deepseek2025r1} & 58.9 & 31.9 & 20.1 & 38.9 & 56.6 \\
Qwen3.5-Flash~\citep{qwen35flash2025}    & 90.1 & 57.3 & 44.3 & 100.0 & 99.6 \\
\midrule
\rowcolor{rowbeige}
\multicolumn{6}{l}{\textbf{\textit{(B$^{\flat}$) Sub-cliff regime ($\leq$2B)}}} \\
SmolLM2-1.7B~\citep{allal2025smollm2}     & 40.2 & 20.5 & 13.0 & 10.5 & 34.3 \\
\midrule
\rowcolor{rowbeige}
\multicolumn{6}{l}{\textbf{\textit{(B) Open-weight (7--10B)}}} \\
Llama-3.1-8B~\citep{grattafiori2024llama3}    & 58.3 & 45.0 & 41.5 & 56.7 & 80.4 \\
Qwen2.5-7B~\citep{qwen2024qwen25}      & 56.7 & 46.1 & 33.0 & 46.7 & 80.1 \\
Mistral-7B~\citep{jiang2023mistral7b}      & 57.3 & 26.5 & 19.0 & 54.4 & 71.7 \\
Gemma-2-9B~\citep{gemma2-2024}      & 61.4 & 53.1 & 40.0 & 60.0 & 84.9 \\
Falcon3-10B~\citep{falcon3-2025}     & 53.8 & 5.6  & 1.8  & 61.1 & 87.7 \\
\midrule
\rowcolor{rowbeige}
\multicolumn{6}{l}{\textbf{\textit{(B$'$) Mid-scale open-weight (27--32B)}}} \\
Qwen2.5-32B~\citep{qwen2024qwen25}    & 66.2 & 76.3 & 73.2 & 97.8 & 98.2 \\
Gemma-3-27B-it~\citep{gemma3-2025}    & 83.3 & 48.7 & 40.8 & 45.8 & 84.6 \\
\midrule
\rowcolor{rowbeige}
\multicolumn{6}{l}{\textbf{\textit{(B$''$) Vision-language (text-only ablation)}}} \\
Qwen2.5-VL-3B~\citep{qwen25vl-2025}   & 50.1 & 36.4 & 25.3 & 35.7 & 42.3 \\
\bottomrule
\end{tabular}
\caption{
\label{tab:l4_counterfactual}
\textbf{L4 strict-vs-partial dissociation and L5c counterfactual graph recovery
(English, direct).}
Left: L4 strict pass (\textbf{pass\%}), event precision/recall
(\textbf{evP\%}/\textbf{evR\%}). Right: L5c counterfactual strict pass and
partial-credit graph F1. Falcon3 achieves 53.8\% strict pass but only 5.6\%
event precision; Qwen2.5-32B reaches 97.8/98.2 on L5c, approaching the GPT-4o
ceiling. Both dissociations support reporting strict and partial-credit
metrics jointly.
}
\end{table}

%% file: tables/script_category.tex

\begin{table}[t]
\centering
\scriptsize
\setlength{\tabcolsep}{3pt}
\renewcommand{\arraystretch}{1.15}
\resizebox{\columnwidth}{!}{%
\begin{tabular}{l cccccc c}
\toprule
\textbf{Script category}
& \textbf{L0} & \textbf{L1} & \textbf{L2} & \textbf{L3} & \textbf{L4} & \textbf{L5}
& \textbf{Gap vs en} \\
\midrule
Latin (en/es/de)  & 50.2 & 65.6 & 37.1 & 14.8 & 37.6 & 57.3 & $-$11.3 \\
CJK (zh/ja/ko)    & 53.0 & 63.7 & 30.3 & \textbf{16.8} & 32.2 & 51.7 & $-$10.0 \\
Arabic (ar)       & 29.1 & 43.5 & 22.2 & 11.4 & 21.5 & 39.6 & $-$24.7 \\
Thai (th)         & 39.5 & 50.9 & 28.5 &  9.5 & 26.3 & 44.7 & $-$20.0 \\
\midrule
\rowcolor{rowbeige}
\multicolumn{8}{l}{\textbf{\textit{Cross-level Kendall's $\tau$ (mean over 8 evaluable models)}}} \\
$\tau$(L0,L1) static$\to$static    & \multicolumn{7}{l}{$+$0.50 \quad (high stability)} \\
$\tau$(L0,L3) static$\to$active    & \multicolumn{7}{l}{$+$0.29 \quad (moderate)} \\
$\tau$(L3,L5) active$\to$generative& \multicolumn{7}{l}{$+$0.18 \quad (low)} \\
$\tau$(L4,L5) active$\to$active    & \multicolumn{7}{l}{$+$0.48 \quad (moderate-high)} \\
\bottomrule
\end{tabular}%
}
\caption{
\label{tab:script_category}
\textbf{Cross-language patterns by script category.}
Top: mean pass rate per script category per level (averaged over nine models);
rightmost column = mean gap from English. CJK slightly outperforms Latin at L3
(16.8 vs 14.8); Arabic and Thai trail. Bottom: cross-level Kendall's $\tau$
over the eight languages---adjacent static levels are highly stable ($+0.50$),
the active--generative pair drops to $+0.18$, showing per-language ranking
reorganizes across the static-to-active boundary.
}
\end{table}

%% file: sections/07_conclusion.tex
\section{Conclusion}
\label{sec:conclusion}

The seven findings argue against treating spatial reasoning as a single capability. Failure modes are \emph{structural} (frame confusion, anchor flips, node--edge dissociations) rather than arithmetic, and the Kendall's $\tau$ discontinuity at the static-to-active boundary ($+0.50\!\to\!+0.18$) shows per-language rankings reorganize across tiers. F7 supplies the strongest reading: \emph{across all eight languages}, native speakers collapse to L3$=$0\% while L4 plateaus at $\sim$41\% (std 3.3pp)---a cross-lingual replication that rules out language-specific accounts and points at the modality. The implied mechanism is a verbal working-memory bottleneck \citep{cowan2001magical}, compatible with probing evidence that LLMs encode partial spatial structure internally \citep{li2024probing,vafa2024evaluating}.

\noindent\textbf{Robustness and implications.} The L3 cliff persists across a $\sim$19$\times$ parameter range (1.7B--32B), four frontier closed-source families, and eight languages in both LLMs and humans; two mid-scale controls (Qwen2.5-32B, Gemma-3-27B) establish it as a mid-capability property. The diagnostic profile motivates externalised-state decoding \citep{nye2021scratchpad\ifpreprint,liao2026spamem\fi} for the L3 cliff, constrained generation \citep{geng2024grammar} for F4, and language-balanced tuning \citep{ustun2024aya} for F6. The cross-lingual human cliff reframes pure-text as a shared bottleneck, motivating multimodal NLP; quantitative falsifiers are pre-registered in Section~\ref{sec:limitations}.

%% file: sections/08_limitations.tex
\section*{Limitations}
\label{sec:limitations}

\noindent\textbf{Scope of language coverage.} \mentalmap{}'s eight languages were chosen to span all three Levinsonian frame classes and both LTR/RTL writing systems while keeping each cell at full evaluation scale; broader typological coverage is a natural extension the released harness supports.

\noindent\textbf{Synthetic scene grounding.} \mentalmap{} is built on ProcTHOR scenes with native-speaker-validated multilingual renderings. Procedural generation is a deliberate choice: it enables the orthogonal-axis design and reproducible scaling that real-world scene collection would not support, and the structured-text control isolates spatial reasoning from natural-language quality. Extension to real-world scene descriptions is straightforward through the released harness.

\noindent\textbf{Modality scope and falsifier design.} The benchmark is intentionally pure-text, which the cross-lingual human pilot (F7) shows is itself a meaningful diagnostic regime: L3 collapse to 0\% in all eight languages and L4 language-invariance ($\sim$41\%, std 3.3pp) directly support a modality account of the cliff. A paired scratchpad-augmented condition is the natural next falsifier of this account, and we commit to two extensions in the next release: \textbf{(i)} a scratchpad-augmented evaluation where both models and human participants receive an external notation buffer, predicted by the modality account to lift L3 disproportionately over L0--L2; and \textbf{(ii)} a multimodal variant rendering the same scene as a floorplan image, predicted to lift L3 further by matching representation to task. The benchmark harness is designed to support both probes without rebuilding the underlying scene pool.

\noindent\textbf{Predicted effect sizes for committed extensions.} Grounded in cognitive-load theory \citep{cowan2001magical} and prior structured-prompting work, we register the following quantitative predictions to allow each extension to be properly falsified: \textbf{(i)} scratchpad-augmented L3 is predicted to rise from $\sim$14\% to 35--50\% Overall (lift $+$20 to $+$35pp, by analogy to scratchpad gains on multi-step arithmetic and reasoning of $+$25 to $+$45pp reported by \citet{nye2021scratchpad} and \citet{wei2022chain}), while L0--L2 should stay within $\pm$5pp; the differential lift is the testable modality prediction. \textbf{(ii)} A multimodal floorplan variant is predicted to lift L3 to 40--55\% ($+$25 to $+$40pp), consistent with VLM-vs-text gains on spatial QA of $+$15 to $+$30pp \citep{zhang2024comfort}, with weaker effects expected at L4 dynamic tasks. \textbf{(iii)} A 70B-class open-weight scale control is predicted to retain the L3 cliff at 15--22\% (a $+$5 to $+$10pp lift over Qwen2.5-32B but well below the $L3\!=\!L0/2$ threshold, consistent with the scale-resistance of spatial reasoning reported by \citet{vafa2024evaluating}), while pushing L5 to 78--85\% (continued open-source/closed-source convergence). These ranges are pre-registered targets, not observed measurements.

\noindent\textbf{Reproducibility and release.} Reproducing the headline findings depends on non-trivial L5 schema normalisers, event-set F1 implementation, and hallucination decomposition logic. We commit to a versioned public release including: rendered items in all eight natural languages and the structured-text control; schema validators and normalisers for L4 event records and L5 graphs; the evaluation harness with deterministic seeds matching the reported numbers; prompt templates for direct and chain-of-thought regimes; the human-pilot terminal harness; and a consistency-check script that recomputes all main-text composites from released traces.

\section*{Ethical Considerations}

\mentalmap{} uses synthetic ProcTHOR scenes with no human subjects or personal data. Native-speaker validation and the human pilot were conducted by paid annotators at standard research-assistant rates with informed consent. The benchmark is released under an open license; we discourage use of aggregate \mentalmap{} scores as the sole criterion for deploying models in safety-critical settings, as the diagnostic axes reveal failure modes that single-number accuracy hides.

%% file: sections/A_appendix.tex
\appendix
\FloatBarrier
\section{Benchmark Design Details}
\label{sec:app_design_details}

\subsection{Dataset Statistics}
\label{sec:app_dataset_stats}

\mentalmap{} is built on 100 ProcTHOR houses (median 5--7 rooms) paired with sampled AI2-THOR action episodes. Each house contributes 36 base tasks organized into the six capability levels, rendered in nine conditions (eight natural languages plus a structured-text control) under two prompt regimes (direct and chain-of-thought). Table~\ref{tab:app_dataset_counts} reports per-level case counts.

\begin{table}[h]
\centering
\scriptsize
\setlength{\tabcolsep}{3pt}
\renewcommand{\arraystretch}{1.1}
\resizebox{\columnwidth}{!}{%
\begin{tabular}{l r r r r}
\toprule
\multicolumn{5}{l}{\textbf{Dataset:} 100 ProcTHOR houses $\times$ 8 natural languages} \\
\multicolumn{5}{l}{$+$ structured-text control $\times$ 3 description modes $\times$ 2 prompt regimes} \\
\midrule
\textbf{Level} & \textbf{Tasks} & \textbf{Cells} & \textbf{Items/lang} & \textbf{Items} \\
\midrule
L0 atomic facts             & 5  & 250  &   750 &  6{,}750 \\
L1 anchor-local relations   & 5  & 250  &   833 &  7{,}500 \\
L2 multi-object integration & 6  & 300  & 1{,}167 & 10{,}500 \\
L3 viewpoint \& direction   & 5  & 250  &   833 &  7{,}500 \\
L4 dynamic state update     & 13 & 650  & 1{,}200 & 10{,}800 \\
L5 generative world-graph   & 5  & 250  &   500 &  4{,}500 \\
\midrule
\textbf{Total}              & \textbf{39} & \textbf{1{,}950} & \textbf{5{,}283} & \textbf{47{,}550} \\
\bottomrule
\end{tabular}%
}

\vspace{6pt}
\begin{center}
\begin{tikzpicture}[scale=0.85]
\fill[paperbeigelight] (0,0) -- (46.15:1.4) arc[start angle=46.15, end angle=0, radius=1.4] -- cycle;
\fill[paperbeige] (0,0) -- (92.31:1.4) arc[start angle=92.31, end angle=46.15, radius=1.4] -- cycle;
\fill[paperlavenderlight] (0,0) -- (147.69:1.4) arc[start angle=147.69, end angle=92.31, radius=1.4] -- cycle;
\fill[paperlavender] (0,0) -- (193.85:1.4) arc[start angle=193.85, end angle=147.69, radius=1.4] -- cycle;
\fill[papernavy] (0,0) -- (313.85:1.4) arc[start angle=313.85, end angle=193.85, radius=1.4] -- cycle;
\fill[paperpink] (0,0) -- (360:1.4) arc[start angle=360, end angle=313.85, radius=1.4] -- cycle;

\node[font=\scriptsize\bfseries] at (23:1.85)   {L0 (5, 12.8\%)};
\node[font=\scriptsize\bfseries] at (69:1.85)   {L1 (5, 12.8\%)};
\node[font=\scriptsize\bfseries] at (120:1.85)  {L2 (6, 15.4\%)};
\node[font=\scriptsize\bfseries] at (171:1.85)  {L3 (5, 12.8\%)};
\node[font=\scriptsize\bfseries, color=papernavy] at (254:1.85)  {\textbf{L4 (13, 33.3\%)}};
\node[font=\scriptsize\bfseries] at (337:1.85)  {L5 (5, 12.8\%)};
\end{tikzpicture}\\[2pt]
{\scriptsize Task-family distribution across levels (39 total). L4 dominates with 13 sub-tasks covering dynamic-state and counterfactual reasoning.}
\end{center}

\caption{
\label{tab:app_dataset_counts}
\textbf{Per-level benchmark composition.} The benchmark is grounded in 100
ProcTHOR houses and rendered in 9 conditions (8 natural languages
$+$ structured-text control), each under 3 description modes (balanced,
count-first, relation-first) and 2 prompt regimes (direct, reasoning).
\emph{Tasks} = unique task families per level. \emph{Cells} = (task $\times$
language $\times$ mode $\times$ regime) evaluation cells.
\emph{Items/lang} = scored items per language condition (across 3 modes $\times$
2 regimes). \emph{Items} = total scored items per model evaluation
(15--60 items per cell). Each of the eleven evaluated model panels in
Sections~\ref{sec:res_staircase}--\ref{sec:res_human_pilot} contributes
47{,}550 items, for $\sim$523{,}000 total evaluation runs.
}
\end{table}

\subsection{Task-Level Metrics}
\label{sec:app_metrics}

\noindent\textbf{L0--L4 metrics.} L0 uses boolean accuracy, count accuracy, and set/histogram F1. L1 uses receptacle-exact match, boolean accuracy, and set F1. L2 uses direction-relaxed accuracy, chain consistency, and indirect-search F1. L3 uses direction-exact accuracy, frame-shift drop, and global$\leftrightarrow$view accuracy. L4 uses receptacle-exact and counterfactual delta F1 (L4a), event-set F1 and attribute-exact match (L4b), and final-state exact match with Kendall's $\tau$ for action ordering (L4c). For each evaluator, an item is marked correct after canonical normalisation (lowercase, whitespace strip, language-specific alias resolution); the cell-level \texttt{passRate} reported in Table~\ref{tab:main_results} is the per-cell binary correctness rate.

\noindent\textbf{L5 composite score.} L5 graphs are scored with a partial-credit composite designed to reflect the multiple ways structured-output predictions can be partially correct. For each predicted graph $\hat{G} = (\hat{V}, \hat{E})$ against gold $G^{*} = (V^{*}, E^{*})$, the evaluator first computes:
\begin{itemize}
\item \emph{JSON validity} ($v_{\textsc{json}} \in \{0,1\}$): whether the output parses as JSON with \texttt{nodes} and \texttt{edges} list fields.
\item \emph{Schema validity} ($v_{\textsc{sch}} \in \{0,1\}$): JSON valid \emph{and} every edge endpoint appears in the declared node set.
\item \emph{Node F1} ($f_{V}$): standard set F1 over \{predicted, gold\} node-type sets, treating each edge endpoint as an implicit node.
\item \emph{Edge F1} ($f_{E}$): set F1 over \{predicted, gold\} edge tuples $(\textit{source}, \textit{target}, \textit{relation})$ after lowercase normalisation.
\item \emph{Relation correctness} ($r$): fraction of predicted edges whose relation label is the canonical \texttt{containedin}; defined as $1$ when both predicted and gold edge sets are empty.
\end{itemize}
The per-item L5 score is then
\begin{equation*}
s_{\text{L5}} = 0.05\,v_{\textsc{json}} + 0.05\,v_{\textsc{sch}} + 0.20\,f_{V} + 0.55\,f_{E} + 0.15\,r,
\end{equation*}
clipped to $[0,1]$. The dominant 0.55 weight on edge F1 reflects that scoring object \emph{containment relations} is the substantive structured-output task; node identification, validity, and relation-type checks act as supporting components. Table~\ref{tab:main_results}'s L5 column reports the cell-level mean of $s_{\text{L5}}$ (\texttt{averageScore}). A separate strict binary flag, \texttt{passed} $= \mathbf{1}\{s_{\text{L5}} \ge 0.90 \land f_{E} \ge 0.85 \land r \ge 0.90\}$, is reported as \texttt{passRate} in supplementary diagnostic tables. We deliberately report both: the strict flag rewards full reconstruction (and is near-zero for most open-weight cells), while the partial-credit composite exposes the node--edge dissociation discussed in Section~\ref{sec:res_metric}.

\noindent\textbf{Cell aggregation.} A \emph{cell} is one (model, level, task, language, descriptionMode, promptRegime) combination, with $n\!=\!15$ items by default. Per-cell scores are aggregated to per-(level, language) composites by equal-weight macro averaging over tasks (filtering to $n\!\ge\!10$ to suppress small-sample noise); the Overall column averages over the eight natural languages with equal weight, excluding the structured-text control.

\noindent\textbf{Diagnostic-axis metrics.} FoR uses per-frame accuracy and frame-shift drop; RTL uses anchor distribution on free-form completions; reasoning uses paired direct/reasoning prompt delta; hallucination uses hallucinated/missing node and edge counts plus node and edge F1.

\subsection{Evaluation Protocol}
\label{sec:app_eval_protocol}

For each task we evaluate two prompting strategies: direct (``answer the question'') and reasoning (``think step by step before answering''). For diagnostic axes we use specialized protocols: frame-of-reference variants are presented zero-shot to avoid in-context FoR contamination; RTL bias is measured on free-form generation rather than multiple choice; reasoning effort is measured via paired direct vs.\ chain-of-thought prompting at every (task, language) cell; hallucination detection uses node/edge F1 against the ProcTHOR ground-truth scene graph at L5 and free-form scene-listing comparison elsewhere.

All evaluations run on standardized infrastructure with deterministic decoding (temperature 0). For closed-source models we use official APIs; for open-weight models we use vLLM~\citep{vllm2023} on a single A100 80GB node. Total evaluation cost was approximately 2{,}400 GPU-hours for open-weight models plus API costs for the three closed-source systems.

\subsection{Implementation Pipeline}
\label{sec:app_pipeline}

The companion code release implements an end-to-end pipeline that consumes three input sources (ProcTHOR house layouts, AI2-THOR action episode traces, and per-language resource files) and produces evaluator-ready result entries organized by (model, case, level, language) keys. Case generation produces two case types: \emph{single-action} cases supporting all six capability levels, and \emph{multi-action} cases for three-action non-overlapping chains restricted to L4. Each case is annotated with axis metadata that maps each task to one or more diagnostic axes, enabling reproducible re-stratification of stored results without re-running models. The L5 graph evaluator automatically computes node and edge hallucination rates as a byproduct of F1 scoring.

\subsection{Annotation and Validation Protocol}
\label{sec:app_annotation}

Per-language description templates were drafted in English and translated by professional translators with linguistics backgrounds into the seven non-English target languages. Each translation was validated by a separate native-speaker reviewer for naturalness and spatial-relation accuracy. The RTL anchor parser was developed against an English-baseline reference set with 200 manually-anchored sentences; we report cross-lingual anchor classification conservatively as a behavioral proxy. Frame-of-reference prompt templates use the canonical three-frame examples from \citet{levinson2003space} as anchors.

\subsection{Frame-of-Reference Prompt Templates}
\label{sec:app_for_templates}

We instantiate three frame-of-reference variants per spatial relation, following the typology of \citet{levinson2003space} and the operationalization of \citet{premsri2025forest}. Table~\ref{tab:app_for_examples} gives an English example for each frame; the same scene is rendered in the eight evaluation languages.

\begin{table}[h]
\centering
\scriptsize
\setlength{\tabcolsep}{4pt}
\renewcommand{\arraystretch}{1.15}
\begin{tabular}{l p{0.62\columnwidth}}
\toprule
\textbf{Frame} & \textbf{English example rendering} \\
\midrule
Relative   & ``The cup is to \emph{my left} on the kitchen counter.'' \\
Absolute   & ``The cup is to the \emph{north} on the kitchen counter.'' \\
Intrinsic  & ``The cup is in \emph{front of} the toaster on the counter.'' \\
\bottomrule
\end{tabular}
\caption{
\label{tab:app_for_examples}
\textbf{Frame-of-reference variants instantiated per spatial relation.}
Each base relation in the ProcTHOR scene graph is rendered in all three frames; FoR variants pair to enable within-model frame-shift analysis.
}
\end{table}

\subsection{Hallucination Evaluator Details}
\label{sec:app_hallucination_detect}

The L5 hallucination evaluator computes four decomposed sub-metrics: (i) \emph{node F1}---token-level F1 over the set of objects in the generated graph vs.\ the ProcTHOR ground-truth scene graph; (ii) \emph{edge F1}---token-level F1 over the set of \texttt{(object, parent receptacle)} containment relations; (iii) \emph{hallucinated-node count}---the number of generated objects not present in the ground-truth graph; (iv) \emph{missing-node count}---the number of ground-truth objects absent from the generated graph. Free-form scene-listing at non-L5 levels uses fuzzy string matching against the canonical ProcTHOR object lexicon (Levenshtein distance threshold 0.85) to identify phantom and missing entities.

\section{Additional Results}
\label{sec:app_additional_results}

\subsection{Per-Model Detailed Results}
\label{sec:app_per_model_results}

Table~\ref{tab:app_per_model_results} expands the main-paper results (Table~\ref{tab:main_results}) into a full per-model breakdown, with one row per condition (eight evaluation languages plus the structured-text control) and one column per level (L0--L5). The structured-text condition serves as a baseline that isolates spatial-reasoning ability from natural-language processing. Table~\ref{tab:app_l0_l2} additionally reports the L0--L2 static-comprehension levels in the same compact \texttt{L0\,/\,L1\,/\,L2} format as the main results table, for direct cross-reference with the L3--L5 view in the main paper.

\input{tables/appendix_per_model}

\input{tables/appendix_l0_l2}

\subsection{L3 Sub-Task and L5 Graph Breakdowns}
\label{sec:app_subtask}

Tables~\ref{tab:app_l3_subtask_lang} and~\ref{tab:app_l5_graph_diagnostics} expand the L3 sub-task and L5 graph diagnostics into full per-language and per-sub-task breakdowns, supporting the sub-task-structured finding (\S\ref{sec:res_l3_cliff}) and the node--edge dissociation finding (\S\ref{sec:res_metric}).

\input{tables/appendix_l3_subtask}

\input{tables/appendix_l5_graph}

\subsection{JSON Validity Under Reasoning Prompts}
\label{sec:app_json_validity}

Table~\ref{tab:app_json_validity} reports the L5 JSON validity rate under direct vs.\ reasoning prompts for all evaluated models, supporting the F3 mechanism that reasoning preambles disrupt strict structured output. Ten of eleven models show degraded JSON validity under reasoning, with Mistral-7B collapsing most severely (70.4\%$\to$7.3\%, $-63.0$pp). The single exception, DeepSeek-V4-Flash ($+5.2$pp), is also the model whose L3 reasoning delta is most positive ($+32.4$pp), suggesting its decoding stack allocates output budget differently from the other systems.

\input{tables/appendix_json_validity}

\subsection{RTL Anchor-Preference Probe}
\label{sec:app_anchor}

Table~\ref{tab:app_anchor_preference} reports per-model anchor-selection rates on the \texttt{rtlAnchorPreference} task, isolating the reading-direction-bias diagnostic axis. In LTR languages all evaluated models prefer the target anchor over the source anchor by 19--34pp; the Arabic and Thai cells fall into the \emph{unknown} bucket, indicating the scope of the current anchor-extraction protocol on non-Latin scripts. Cross-language anchor preference under a multilingually-calibrated extractor is a natural follow-up extension to this diagnostic axis.

\input{tables/appendix_anchor_preference}

\subsection{Supplementary Visualizations}
\label{sec:app_supp_figures}

\begin{figure}[t]
\centering
\includegraphics[width=\linewidth]{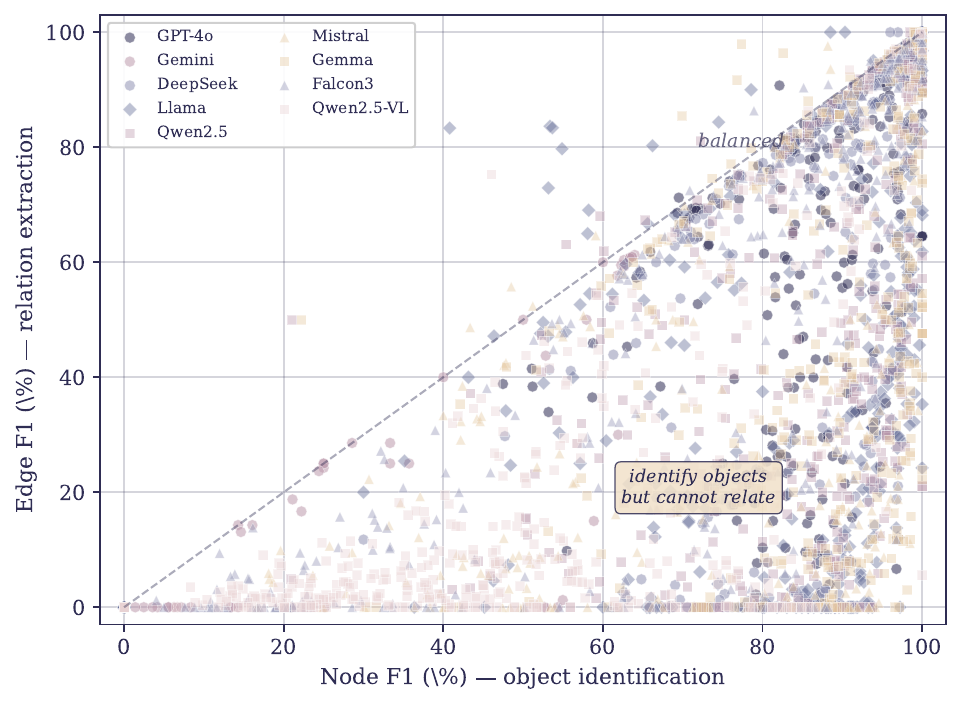}
\caption{
\label{fig:app_node_edge}
\textbf{Full L5 node F1 vs.\ edge F1 scatter} across all (model, sub-task, language) cells. Diagonal marks $node\!=\!edge$. Marker shape distinguishes closed-source ($\bullet$) from open-weight models.
}
\end{figure}

\begin{figure}[t]
\centering
\includegraphics[width=\linewidth]{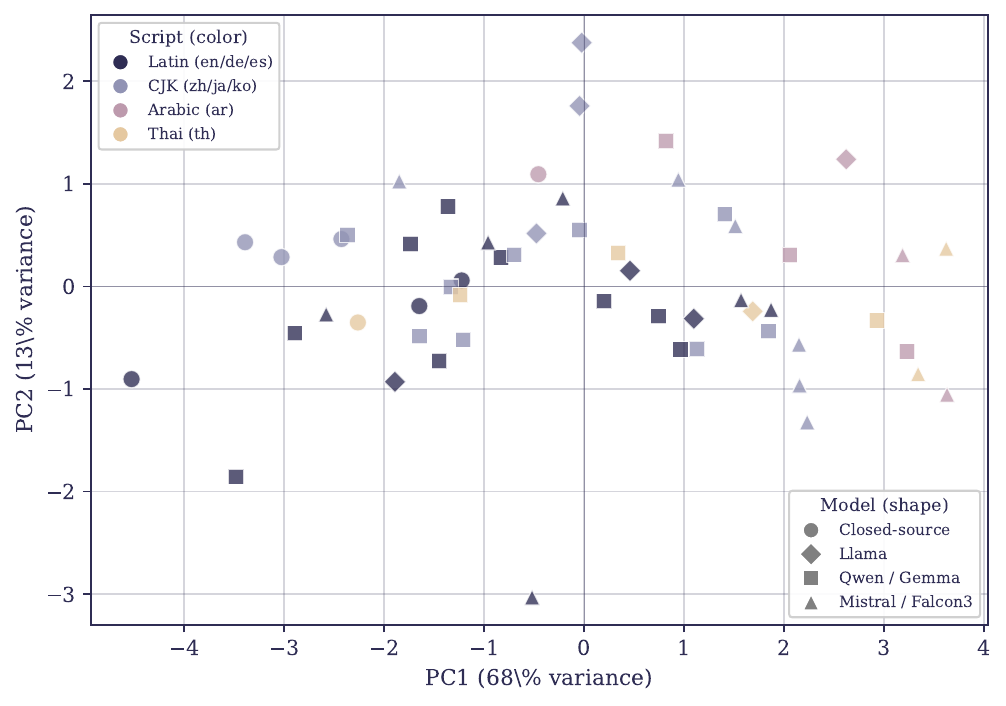}
\caption{
\label{fig:app_pca}
\textbf{PCA embedding of (model, language) profiles} over L0--L5. Color encodes script family; shape encodes model. PC1 captures a multilingual-coverage gradient.
}
\end{figure}

Figures~\ref{fig:app_node_edge} and~\ref{fig:app_pca} present two complementary views of the data. Figure~\ref{fig:app_node_edge} shows the full per-(model, sub-task, language) scatter of L5 node F1 against edge F1. The lower triangle is densely populated while the upper triangle is sparse, indicating that node identification universally exceeds relation extraction across all models, languages, and L5 sub-tasks. Outliers on the upper-left (high edge F1 with low node F1) are rare, occurring mainly for cells where the model produced few but correctly-related entities. Closed-source models (circles) populate the upper-right corner with both high node and edge F1; open-weight 7--10B models (squares, triangles, diamonds) are more spread, particularly along the node axis at low edge F1 values---visualizing the F4 dissociation finding in full granularity.

Figure~\ref{fig:app_pca} shows a PCA embedding of the per-(model, language) profile vectors across L0--L5 capability levels. PC1 (68\% of variance) corresponds to a multilingual-coverage gradient: Latin-script points (circles) cluster on the left of the embedding, while Arabic and Thai points (squares, diamonds) cluster on the right; CJK points (triangles) populate the middle. PC2 (13\% of variance) does not admit a clean linguistic interpretation. The embedding corroborates the dendrogram-based clustering in Figure~\ref{fig:multilingual_fingerprint} from an independent dimensionality-reduction perspective: the same script-coverage gradient that the hierarchical clustering recovers as a discrete typology also emerges as the dominant continuous axis under PCA. Together, the two visualizations triangulate the F6 finding that multilingual breadth is a structural property of the (model, language) interaction.

\section{Human Pilot Protocol}
\label{sec:app_human_pilot}

This appendix documents the human-pilot protocol underlying F7 (\S\ref{sec:res_human_pilot}).

\noindent\textbf{Participants.} Native speakers of each of the eight evaluation languages (en, zh, ja, ko, es, ar, th, de), N$\geq$3 participants per language, all university-educated, none of whom are authors of this work and none previously exposed to the ProcTHOR scene pool. Compensation was at standard research-assistant rate. Informed consent was obtained.

\noindent\textbf{Item sample.} A stratified random sample drawn from each language split of the main evaluation pool, covering L3 (viewpoint and global direction) and L4 (dynamic state update) cells. The L3 sample covers \texttt{rotationCheck}, \texttt{viewToGlobal}, and \texttt{globalToView}; the L4 sample covers \texttt{event}, \texttt{undo}, \texttt{counterfactualLocation}, \texttt{partialObservation}, \texttt{eventSet}, \texttt{actionFailure}, \texttt{multiEvent}, and \texttt{beliefRevision}. L5 generative-graph output was excluded because emitting a valid JSON scene graph without machine aids is not a meaningful human task. The structured-text control was also excluded. The sample is intentionally biased toward harder items to probe the cliff region.

\noindent\textbf{Difficulty tiers.} Each item was a priori tagged \textsc{Easy} / \textsc{Medium} / \textsc{Hard} from two structural features of its prompt, scored before participant attempt: (i) \emph{referent count}---the number of distinct objects mentioned in the scene description ($<\!8$ Easy, $8$--$15$ Medium, $>\!15$ Hard); (ii) \emph{required chain depth}---the number of anchor relations that must be composed to reach an answer ($1$ Easy, $2$ Medium, $\ge 3$ Hard). The maximum of the two scores determined the tier.

\noindent\textbf{Presentation.} A terminal-based harness rendered each item in three blocks---scene description, question, free-form answer field---identical in content to the LLM prompt. Participants typed an answer or typed \texttt{skip}; no scratchpad, no diagram, no re-reading of earlier items, and no external aids were permitted. There was no time limit; participants chose when to advance.

\noindent\textbf{Scoring.} Strict, identical to the LLM main-evaluation pipeline: skips count as $0$. L3 tasks use direction-exact accuracy and global$\leftrightarrow$view accuracy; L4 tasks use the receptacle-exact / event-set F1 / final-state metrics described in Appendix~\ref{sec:app_metrics}.

\noindent\textbf{Cross-lingual results.} L3 collapses to 0\% in all eight languages; L4 is language-invariant, ranging from 35.4\% (Thai) to 45.2\% (Japanese) with mean 41.1\% and std 3.3pp---no language deviates from the cross-language L4 mean by more than 6pp. The identical L3 collapse and tightly-clustered L4 ceiling across script families spanning all three Levinsonian frame-of-reference systems is direct evidence that the cliff is a property of \emph{pure-text presentation}; the paired scratchpad extension committed in Section~\ref{sec:limitations} is the next falsifier.

%% file: tables/appendix_per_model.tex

\begin{table*}[!htbp]
\centering
\scriptsize
\setlength{\tabcolsep}{3.5pt}
\renewcommand{\arraystretch}{1.05}
\begin{minipage}[t]{0.48\linewidth}
\centering
\begin{tabular}{l cccccc}
\toprule
\textbf{Lang} & \textbf{L0} & \textbf{L1} & \textbf{L2} & \textbf{L3} & \textbf{L4} & \textbf{L5}$^\dagger$ \\
\midrule
\rowcolor{rowlavender}
\multicolumn{7}{l}{\textbf{\textit{GPT-4o~\citep{openai2024gpt4o}}}} \\
en      & -- & -- & -- & 32.4 & 52.4 & 93.3 \\
zh      & -- & -- & -- & 30.9 & 44.3 & 75.8 \\
ja      & -- & -- & -- & 24.2 & 40.5 & 77.6 \\
ko      & -- & -- & -- & 36.3 & 40.8 & 70.7 \\
es      & -- & -- & -- & 23.3 & 39.2 & 81.9 \\
ar      & -- & -- & -- & 36.0 & 29.8 & 55.1 \\
th      & -- & -- & -- & 26.3 & 37.8 & 71.6 \\
de      & -- & -- & -- & 24.8 & 35.9 & 69.1 \\
Struct. & -- & -- & -- & 34.2 & 61.5 & 99.4 \\
\midrule
\rowcolor{rowlavender}
\multicolumn{7}{l}{\textbf{\textit{Gemini-2.5-Flash~\citep{google2025gemini25}}}} \\
en      & -- & -- & -- & 21.3 & 61.2 & 31.6 \\
zh      & -- & -- & -- & 17.5 & 47.9 & 9.2 \\
ja      & -- & -- & -- & 15.1 & 7.6  & 9.7 \\
ko      & -- & -- & -- & 1.3  & 10.2 & 8.0 \\
es      & -- & -- & -- & 4.2  & 13.0 & 10.0 \\
ar      & -- & -- & -- & 2.4  & 3.6  & 6.9 \\
th      & -- & -- & -- & 2.6  & 8.4  & 7.5 \\
de      & -- & -- & -- & 5.2  & 4.0  & 8.0 \\
Struct. & -- & -- & -- & 29.3 & 49.3 & 38.7 \\
\midrule
\rowcolor{rowlavender}
\multicolumn{7}{l}{\textbf{\textit{DeepSeek-V4-Flash~\citep{deepseek2025r1}}}} \\
en      & 89.2 & 82.7 & 52.3 & 26.4 & 56.9 & 71.0 \\
zh      & 84.0 & 90.5 & 39.1 & 27.8 & 45.4 & 66.2 \\
ja      & 89.2 & 93.2 & 37.6 & 23.1 & 40.5 & 64.3 \\
ko      & 72.4 & 94.9 & 34.0 & 20.7 & 40.3 & 65.7 \\
es      & 67.2 & 72.2 & 30.0 & 15.9 & 45.0 & 65.4 \\
ar      & 45.5 & 74.5 & 31.5 & 23.1 & 29.8 & 51.1 \\
th      & 70.8 & 86.0 & 36.0 & 17.0 & 45.4 & 62.0 \\
de      & 54.5 & 81.7 & 31.1 & 16.8 & 41.5 & 56.4 \\
Struct. & 94.7 & 86.9 & 47.8 & 22.0 & 66.3 & 72.6 \\
\midrule
\rowcolor{rowlavender}
\multicolumn{7}{l}{\textbf{\textit{Qwen3.5-Flash~\citep{qwen35flash2025}}}} \\
en      & 84.7 & 91.8 & 52.0 & 6.2  & 76.4 & 98.7 \\
zh      & 85.4 & 98.1 & 49.0 & 12.5 & 66.9 & 90.9 \\
ja      & 85.2 & 96.5 & 49.9 & 14.4 & 51.1 & 83.3 \\
ko      & 44.3 & 95.6 & 44.5 & 3.1  & 44.0 & 78.5 \\
es      & 79.7 & 84.4 & 38.8 & 6.9  & 55.7 & 97.8 \\
ar      & 65.8 & 87.5 & 38.1 & 9.4  & 46.2 & 73.1 \\
th      & 82.6 & 94.0 & 42.8 & 13.8 & 48.0 & 96.0 \\
de      & 85.2 & 97.7 & 46.4 & 6.2  & 60.7 & 85.9 \\
Struct. & 99.3 & 95.9 & 52.1 & 13.8 & 77.2 & 97.8 \\
\midrule
\rowcolor{rowbeige}
\multicolumn{7}{l}{\textbf{\textit{SmolLM2-1.7B-Instruct (B$^{\flat}$ sub-cliff regime)~\citep{allal2025smollm2}}}} \\
en      & 36.3 & 18.4 & 42.2 & 0.0  & 37.6 & 41.4 \\
zh      & 35.9 & 24.0 & 17.3 & 4.3  & 27.7 & 4.5  \\
ja      & 23.5 & 17.9 & 22.8 & 0.0  & 26.2 & 14.4 \\
ko      & 37.8 & 11.0 & 15.6 & 0.0  & 26.4 & 5.7  \\
es      & 16.8 & 12.5 & 17.0 & 0.0  & 25.4 & 22.4 \\
ar      & 27.5 & 30.0 & 22.7 & 20.0 & 20.5 & 8.7  \\
th      & 34.9 & 30.0 & 21.8 & 20.0 & 18.1 & 3.0  \\
de      & 23.4 & 14.3 & 16.2 & 0.0  & 23.8 & 20.3 \\
Struct. & 40.7 & 27.5 & 46.1 & 0.9  & 43.1 & 26.2 \\
\midrule
\rowcolor{rowbeige}
\multicolumn{7}{l}{\textbf{\textit{Llama-3.1-8B-Instruct~\citep{grattafiori2024llama3}}}} \\
en      & 62.7 & 74.7 & 41.5 & 13.3 & 45.1 & 62.0 \\
zh      & 38.3 & 73.0 & 31.3 & 16.6 & 35.2 & 60.8 \\
ja      & 37.5 & 45.1 & 28.1 & 30.2 & 30.5 & 59.0 \\
ko      & 41.4 & 65.5 & 20.6 & 29.4 & 26.7 & 58.5 \\
es      & 36.0 & 50.4 & 31.2 & 10.8 & 30.1 & 64.2 \\
ar      & 25.4 & 31.1 & 18.9 & 14.2 & 16.7 & 46.2 \\
th      & 30.0 & 36.3 & 22.3 & 6.9  & 32.6 & 51.5 \\
de      & 24.7 & 59.8 & 28.3 & 7.1  & 33.4 & 48.2 \\
Struct. & 64.7 & 70.4 & 43.2 & 13.3 & 52.0 & 76.6 \\
\midrule
\rowcolor{rowbeige}
\multicolumn{7}{l}{\textbf{\textit{Qwen2.5-7B-Instruct~\citep{qwen2024qwen25}}}} \\
en      & 76.4 & 74.6 & 48.5 & 18.9 & 41.0 & 73.8 \\
zh      & 62.8 & 75.3 & 39.8 & 12.7 & 35.5 & 56.3 \\
ja      & 63.7 & 60.4 & 35.9 & 15.1 & 27.1 & 65.2 \\
ko      & 54.1 & 50.4 & 29.5 & 14.8 & 28.0 & 66.7 \\
es      & 59.5 & 57.9 & 31.9 & 17.7 & 35.3 & 64.0 \\
ar      & 34.2 & 46.6 & 22.8 & 5.9  & 19.9 & 48.6 \\
th      & 44.0 & 57.0 & 30.5 & 9.9  & 26.1 & 65.6 \\
de      & 36.2 & 57.6 & 33.4 & 11.1 & 33.5 & 58.1 \\
Struct. & 94.7 & 67.6 & 40.2 & 16.0 & 55.8 & 83.8 \\
\bottomrule
\end{tabular}
\end{minipage}%
\hfill
\begin{minipage}[t]{0.48\linewidth}
\centering
\begin{tabular}{l cccccc}
\toprule
\textbf{Lang} & \textbf{L0} & \textbf{L1} & \textbf{L2} & \textbf{L3} & \textbf{L4} & \textbf{L5}$^\dagger$ \\
\midrule
\rowcolor{rowbeige}
\multicolumn{7}{l}{\textbf{\textit{Mistral-7B-Instruct-v0.3~\citep{jiang2023mistral7b}}}} \\
en      & 51.1 & 47.6 & 44.0 & 1.0  & 53.5 & 44.2 \\
zh      & 30.9 & 26.5 & 27.8 & 5.7  & 35.3 & 26.8 \\
ja      & 50.0 & 34.7 & 23.9 & 5.3  & 28.3 & 25.1 \\
ko      & 36.5 & 34.7 & 22.8 & 8.3  & 30.9 & 29.8 \\
es      & 24.3 & 36.6 & 30.6 & 14.8 & 31.7 & 36.3 \\
ar      & 19.0 & 25.2 & 26.0 & 0.8  & 22.6 & 19.2 \\
th      & 31.4 & 22.8 & 26.7 & 4.0  & 20.3 & 18.5 \\
de      & 16.6 & 33.9 & 32.2 & 14.7 & 31.4 & 32.3 \\
Struct. & 70.0 & 42.0 & 50.6 & 2.0  & 53.4 & 57.2 \\
\midrule
\rowcolor{rowbeige}
\multicolumn{7}{l}{\textbf{\textit{Gemma-2-9B-it~\citep{gemma2-2024}}}} \\
en      & 69.7 & 80.4 & 61.9 & 13.7 & 48.9 & 70.1 \\
zh      & 60.3 & 87.8 & 39.4 & 13.0 & 39.4 & 58.4 \\
ja      & 56.6 & 88.4 & 44.2 & 26.4 & 39.3 & 58.9 \\
ko      & 49.7 & 86.7 & 42.8 & 16.6 & 33.8 & 55.6 \\
es      & 50.7 & 79.8 & 40.8 & 24.7 & 39.4 & 55.6 \\
ar      & 32.9 & 69.7 & 17.7 & 14.3 & 26.2 & 61.2 \\
th      & 49.2 & 83.0 & 42.8 & 14.1 & 32.7 & 61.2 \\
de      & 42.4 & 81.4 & 39.7 & 25.9 & 36.2 & 53.2 \\
Struct. & 92.9 & 81.3 & 57.5 & 19.7 & 65.7 & 90.4 \\
\midrule
\rowcolor{rowbeige}
\multicolumn{7}{l}{\textbf{\textit{Falcon3-10B-Instruct~\citep{falcon3-2025}}}} \\
en      & 63.5 & 74.6 & 42.0 & 16.4 & 44.5 & 82.0 \\
zh      & 60.0 & 85.6 & 29.2 & 19.4 & 36.8 & 78.4 \\
ja      & 32.9 & 56.0 & 24.0 & 14.6 & 25.0 & 58.9 \\
ko      & 42.7 & 48.9 & 23.2 & 11.0 & 21.2 & 48.3 \\
es      & 52.1 & 74.0 & 33.4 & 14.6 & 32.7 & 69.7 \\
ar      & 16.0 & 39.7 & 22.2 & 7.8  & 18.3 & 29.2 \\
th      & 20.7 & 39.4 & 21.3 & 7.1  & 12.7 & 21.3 \\
de      & 27.3 & 74.6 & 30.8 & 14.4 & 31.4 & 70.4 \\
Struct. & 99.3 & 83.9 & 35.8 & 4.0  & 56.4 & 82.8 \\
\midrule
\rowcolor{rowbeige}
\multicolumn{7}{l}{\textbf{\textit{Qwen2.5-32B-Instruct (B$'$ scale control)~\citep{qwen2024qwen25}}}} \\
en      & 86.8 & 87.6 & 48.8 & 14.7 & 56.2 & 95.7 \\
zh      & 90.7 & 86.2 & 43.2 & 23.3 & 47.3 & 80.7 \\
ja      & 73.4 & 88.8 & 37.3 & 8.9  & 36.9 & 61.3 \\
ko      & 67.6 & 91.6 & 33.8 & 6.6  & 37.6 & 61.2 \\
es      & 63.4 & 76.7 & 37.7 & 17.7 & 40.7 & 68.5 \\
ar      & 47.2 & 68.6 & 30.2 & 10.2 & 25.3 & 59.8 \\
th      & 68.3 & 85.6 & 40.3 & 18.0 & 40.9 & 72.3 \\
de      & 52.3 & 77.3 & 29.0 & 9.9  & 39.6 & 73.4 \\
Struct. & 100.0 & 80.4 & 51.0 & 9.7  & 68.7 & 89.3 \\
\midrule
\rowcolor{rowbeige}
\multicolumn{7}{l}{\textbf{\textit{Gemma-3-27B-it (B$'$ scale control)~\citep{gemma3-2025}}}} \\
en      & 80.7 & 98.9 & 42.8 & 28.7 & 70.6 & 87.5 \\
zh      & 82.5 & 98.7 & 39.8 & 17.3 & 68.7 & 77.6 \\
ja      & 80.9 & 95.3 & 44.6 & 28.7 & 60.8 & 81.1 \\
ko      & 78.5 & 98.9 & 35.3 & 35.3 & 57.8 & 81.3 \\
es      & 76.5 & 88.0 & 35.2 & 18.0 & 65.3 & 97.2 \\
ar      & 67.2 & 94.9 & 28.1 & 26.0 & 49.3 & 75.4 \\
th      & 77.3 & 95.8 & 37.6 & 10.7 & 57.7 & 93.8 \\
de      & 82.0 & 95.5 & 27.5 & 14.7 & 59.8 & 78.4 \\
Struct. & 100.0 & 93.3 & 39.4 & 23.3 & 73.2 & 99.0 \\
\midrule
\rowcolor{rowbeige}
\multicolumn{7}{l}{\textbf{\textit{Qwen2.5-VL-3B-Instruct~\citep{qwen25vl-2025}}}} \\
en      & 58.1 & 70.0 & 43.5 & 11.2 & 37.6 & 66.4 \\
zh      & 54.7 & 60.9 & 21.9 & 7.7  & 28.1 & 50.1 \\
ja      & 49.5 & 38.9 & 19.8 & 1.8  & 26.1 & 49.6 \\
ko      & 45.6 & 39.5 & 21.3 & 16.9 & 25.9 & 42.6 \\
es      & 44.2 & 49.6 & 27.9 & 7.6  & 31.2 & 55.5 \\
ar      & 30.6 & 17.6 & 16.0 & 0.6  & 26.7 & 39.0 \\
th      & 30.5 & 32.0 & 19.9 & 1.4  & 20.9 & 43.4 \\
de      & 47.0 & 62.9 & 24.6 & 4.1  & 29.5 & 54.8 \\
Struct. & 86.5 & 62.4 & 48.8 & 10.7 & 45.0 & 58.0 \\
\bottomrule
\end{tabular}
\end{minipage}%
\caption{
\label{tab:app_per_model_results}
\textbf{Per-model results across six levels and nine conditions.}
Each model block: nine rows (eight languages + \textbf{Struct.}), six level
columns. L0--L4 use macro-average over tasks ($n\!\ge\!10$); $^\dagger$L5
uses the partial-credit composite (graph F1). Frontier closed-source highlighted
in lavender, open-weight (7--10B) in beige. Dashes mark cells with insufficient
data. 
}
\end{table*}

%% file: tables/appendix_l0_l2.tex

\providecommand{\sep}{\textcolor{black!35}{/}}

\begin{table*}[!htbp]
\centering
\scriptsize
\setlength{\tabcolsep}{1pt}
\renewcommand{\arraystretch}{1.25}
\resizebox{\textwidth}{!}{%
\begin{tabular}{l cccccccc c c}
\toprule
& \multicolumn{8}{c}{\textbf{Language}}
& \multicolumn{2}{c}{} \\
\cmidrule(lr){2-9}
\textbf{Model}
& \textbf{en} & \textbf{zh} & \textbf{ja} & \textbf{ko} & \textbf{es} & \textbf{ar} & \textbf{th} & \textbf{de}
& \textbf{Struct.} & \textbf{Overall} \\
\midrule
\rowcolor{rowlavender}
\multicolumn{11}{l}{\textbf{\textit{(A) Frontier closed-source}}} \\
GPT-4o
  & --\sep--\sep-- & --\sep--\sep-- & --\sep--\sep-- & --\sep--\sep--
  & --\sep--\sep-- & --\sep--\sep-- & --\sep--\sep-- & --\sep--\sep--
  & --\sep--\sep-- & --\sep--\sep-- \\
Gemini-2.5-Flash
  & --\sep--\sep-- & --\sep--\sep-- & --\sep--\sep-- & --\sep--\sep--
  & --\sep--\sep-- & --\sep--\sep-- & --\sep--\sep-- & --\sep--\sep--
  & --\sep--\sep-- & --\sep--\sep-- \\
DeepSeek-V4-Flash
  & 89.2\sep82.7\sep52.3
  & 84.0\sep90.5\sep39.1
  & 89.2\sep93.2\sep37.6
  & 72.4\sep94.9\sep34.0
  & 67.2\sep72.2\sep30.0
  & 45.5\sep74.5\sep31.5
  & 70.8\sep86.0\sep36.0
  & 54.5\sep81.7\sep31.1
  & 94.7\sep86.9\sep47.8
  & 71.6\sep84.5\sep36.5 \\
Qwen3.5-Flash
  & 84.7\sep91.8\sep52.0
  & 85.4\sep98.1\sep49.0
  & 85.2\sep96.5\sep49.9
  & 44.3\sep95.6\sep44.5
  & 79.7\sep84.4\sep38.8
  & 65.8\sep87.5\sep38.1
  & 82.6\sep94.0\sep42.8
  & 85.2\sep97.7\sep46.4
  & 99.3\sep95.9\sep52.1
  & 76.6\sep93.2\sep45.2 \\
\midrule
\rowcolor{rowbeige}
\multicolumn{11}{l}{\textbf{\textit{(B$^{\flat}$) Sub-cliff regime ($\leq$2B)}}} \\
SmolLM2-1.7B-Instruct
  & 36.3\sep18.4\sep42.2
  & 35.9\sep24.0\sep17.3
  & 23.5\sep17.9\sep22.8
  & 37.8\sep11.0\sep15.6
  & 16.8\sep12.5\sep17.0
  & 27.5\sep30.0\sep22.7
  & 34.9\sep30.0\sep21.8
  & 23.4\sep14.3\sep16.2
  & 40.7\sep27.5\sep46.1
  & 29.5\sep19.8\sep21.9 \\
\midrule
\rowcolor{rowbeige}
\multicolumn{11}{l}{\textbf{\textit{(B) Open-weight (7--10B)}}} \\
Llama-3.1-8B-Instruct
  & 62.7\sep74.7\sep41.5
  & 38.3\sep73.0\sep31.3
  & 37.5\sep45.1\sep28.1
  & 41.4\sep65.5\sep20.6
  & 36.0\sep50.4\sep31.2
  & 25.4\sep31.1\sep18.9
  & 30.0\sep36.3\sep22.3
  & 24.7\sep59.8\sep28.3
  & 64.7\sep70.4\sep43.2
  & 37.0\sep54.5\sep27.8 \\
Qwen2.5-7B-Instruct
  & 76.4\sep74.6\sep48.5
  & 62.8\sep75.3\sep39.8
  & 63.7\sep60.4\sep35.9
  & 54.1\sep50.4\sep29.5
  & 59.5\sep57.9\sep31.9
  & 34.2\sep46.6\sep22.8
  & 44.0\sep57.0\sep30.5
  & 36.2\sep57.6\sep33.4
  & 94.7\sep67.6\sep40.2
  & 53.9\sep60.0\sep34.0 \\
Mistral-7B-Instruct-v0.3
  & 51.1\sep47.6\sep44.0
  & 30.9\sep26.5\sep27.8
  & 50.0\sep34.7\sep23.9
  & 36.5\sep34.7\sep22.8
  & 24.3\sep36.6\sep30.6
  & 19.0\sep25.2\sep26.0
  & 31.4\sep22.8\sep26.7
  & 16.6\sep33.9\sep32.2
  & 70.0\sep42.0\sep50.6
  & 32.5\sep32.8\sep29.2 \\
Gemma-2-9B-it
  & 69.7\sep80.4\sep61.9
  & 60.3\sep87.8\sep39.4
  & 56.6\sep88.4\sep44.2
  & 49.7\sep86.7\sep42.8
  & 50.7\sep79.8\sep40.8
  & 32.9\sep69.7\sep17.7
  & 49.2\sep83.0\sep42.8
  & 42.4\sep81.4\sep39.7
  & 92.9\sep81.3\sep57.5
  & 51.4\sep82.2\sep41.2 \\
Falcon3-10B-Instruct
  & 63.5\sep74.6\sep42.0
  & 60.0\sep85.6\sep29.2
  & 32.9\sep56.0\sep24.0
  & 42.7\sep48.9\sep23.2
  & 52.1\sep74.0\sep33.4
  & 16.0\sep39.7\sep22.2
  & 20.7\sep39.4\sep21.3
  & 27.3\sep74.6\sep30.8
  & 99.3\sep83.9\sep35.8
  & 39.4\sep61.6\sep28.3 \\
\midrule
\rowcolor{rowbeige}
\multicolumn{11}{l}{\textbf{\textit{(B$'$) Mid-scale open-weight (27--32B)}}} \\
Qwen2.5-32B-Instruct
  & 86.8\sep87.6\sep48.8
  & 90.7\sep86.2\sep43.2
  & 73.4\sep88.8\sep37.3
  & 67.6\sep91.6\sep33.8
  & 63.4\sep76.7\sep37.7
  & 47.2\sep68.6\sep30.2
  & 68.3\sep85.6\sep40.3
  & 52.3\sep77.3\sep29.0
  & 100.0\sep80.4\sep51.0
  & 68.7\sep82.8\sep37.5 \\
Gemma-3-27B-it
  & 80.7\sep98.9\sep42.8
  & 82.5\sep98.7\sep39.8
  & 80.9\sep95.3\sep44.6
  & 78.5\sep98.9\sep35.3
  & 76.5\sep88.0\sep35.2
  & 67.2\sep94.9\sep28.1
  & 77.3\sep95.8\sep37.6
  & 82.0\sep95.5\sep27.5
  & 100.0\sep93.3\sep39.4
  & 78.2\sep95.7\sep36.4 \\
\midrule
\rowcolor{rowbeige}
\multicolumn{11}{l}{\textbf{\textit{(B$''$) Vision-language (text-only ablation)}}} \\
Qwen2.5-VL-3B-Instruct
  & 58.1\sep70.0\sep43.5
  & 54.7\sep60.9\sep21.9
  & 49.5\sep38.9\sep19.8
  & 45.6\sep39.5\sep21.3
  & 44.2\sep49.6\sep27.9
  & 30.6\sep17.6\sep16.0
  & 30.5\sep32.0\sep19.9
  & 47.0\sep62.9\sep24.6
  & 86.5\sep62.4\sep48.8
  & 45.0\sep46.4\sep24.4 \\
\bottomrule
\end{tabular}%
}
\caption{
\label{tab:app_l0_l2}
\textbf{Static-comprehension results (L0--L2) across eight languages.}
Same layout as main Table~\ref{tab:main_results}, but for the three
static-comprehension levels: L0 (atomic), L1 (anchor), L2 (multi-object).
Each cell shows three composite scores as \texttt{L0\,/\,L1\,/\,L2} (\%).
All cells use the macro-average over tasks with $n\!\ge\!10$ cases (no
partial-credit ambiguity at L0--L2). GPT-4o and Gemini-2.5-Flash did not
run L0--L2 evaluation and are listed for cross-reference.
}
\end{table*}

%% file: tables/appendix_l3_subtask.tex

\begin{table}[!htbp]
\centering
\scriptsize
\setlength{\tabcolsep}{3pt}
\renewcommand{\arraystretch}{1.0}
\resizebox{\columnwidth}{!}{%
\begin{tabular}{l l cccccccc}
\toprule
\textbf{Model} & \textbf{L3 sub-task} & \textbf{en} & \textbf{zh} & \textbf{ja} & \textbf{ko} & \textbf{es} & \textbf{ar} & \textbf{th} & \textbf{de} \\
\midrule
\rowcolor{rowlavender}
\multicolumn{10}{l}{\textbf{\textit{(A) Frontier closed-source}}} \\
\multirow{3}{*}{GPT-4o~\citep{openai2024gpt4o}} & rot & 1.9 & 6.5 & 4.6 & 26.9 & 0.0 & 25.9 & 10.2 & 11.1 \\
 & rotChk & 76.9 & 75.0 & 31.5 & 55.6 & 34.3 & 86.1 & 83.3 & 49.1 \\
 & v$\to$g & 45.4 & 51.9 & 21.3 & 38.0 & 31.5 & 43.5 & 28.7 & 10.2 \\
\midrule
\multirow{3}{*}{Gemini-2.5-Flash~\citep{google2025gemini25}} & rot & 8.3 & 4.2 & 29.2 & 0.0 & 0.0 & 0.0 & 0.0 & 0.0 \\
 & rotChk & 4.2 & 0.0 & 0.0 & 0.0 & 0.0 & 0.0 & 0.0 & 0.0 \\
 & v$\to$g & 56.2 & 56.2 & 22.9 & 2.1 & 0.0 & 0.0 & 0.0 & 0.0 \\
\midrule
\multirow{3}{*}{DeepSeek-V4-Flash~\citep{deepseek2025r1}} & rot & 0.0 & 0.0 & 4.4 & 0.0 & 0.0 & 31.1 & 14.4 & 0.0 \\
 & rotChk & 50.0 & 81.1 & 0.0 & 2.2 & 18.9 & 5.6 & 45.6 & 8.9 \\
 & v$\to$g & 1.1 & 30.0 & 25.6 & 0.0 & 10.0 & 5.6 & 0.0 & 4.4 \\
\midrule
\multirow{3}{*}{Qwen3.5-Flash~\citep{qwen35flash2025}} & rot & 0.0 & 0.0 & 6.2 & 0.0 & 0.0 & 0.0 & 0.0 & 0.0 \\
 & rotChk & 0.0 & 21.9 & 50.0 & 12.5 & 3.1 & 9.4 & 28.1 & 0.0 \\
 & v$\to$g & 31.2 & 40.6 & 9.4 & 3.1 & 31.2 & 37.5 & 40.6 & 31.2 \\
\midrule
\rowcolor{rowbeige}
\multicolumn{10}{l}{\textbf{\textit{(B$^{\flat}$) Sub-cliff regime ($\leq$2B)}}} \\
\multirow{3}{*}{SmolLM2-1.7B~\citep{allal2025smollm2}} & rot & 0.0 & 2.2 & 0.0 & 0.0 & 0.0 & 0.0 & 0.0 & 0.0 \\
 & rotChk & 0.0 & 1.7 & 0.0 & 0.0 & 0.0 & 100.0 & 100.0 & 0.0 \\
 & v$\to$g & 0.0 & 0.0 & 0.0 & 0.0 & 0.0 & 0.0 & 0.0 & 0.0 \\
\midrule
\rowcolor{rowbeige}
\multicolumn{10}{l}{\textbf{\textit{(B) Open-weight (7--10B)}}} \\
\multirow{3}{*}{Llama-3.1-8B~\citep{grattafiori2024llama3}} & rot & 2.2 & 28.9 & 48.9 & 50.0 & 17.8 & 5.6 & 24.4 & 1.1 \\
 & rotChk & 0.0 & 1.1 & 0.0 & 5.6 & 0.0 & 0.0 & 0.0 & 0.0 \\
 & v$\to$g & 28.9 & 0.0 & 0.0 & 0.0 & 0.0 & 3.3 & 6.7 & 0.0 \\
\midrule
\multirow{3}{*}{Qwen2.5-7B~\citep{qwen2024qwen25}} & rot & 25.6 & 36.7 & 33.3 & 42.2 & 31.1 & 0.0 & 17.8 & 18.9 \\
 & rotChk & 90.0 & 3.3 & 0.0 & 0.0 & 3.3 & 16.7 & 0.0 & 0.0 \\
 & v$\to$g & 13.3 & 0.0 & 1.1 & 0.0 & 43.3 & 20.0 & 0.0 & 0.0 \\
\midrule
\multirow{3}{*}{Mistral-7B~\citep{jiang2023mistral7b}} & rot & 0.0 & 1.1 & 3.3 & 2.2 & 10.0 & 1.1 & 0.0 & 0.0 \\
 & rotChk & 7.8 & 0.0 & 7.8 & 14.4 & 33.3 & 2.2 & 10.0 & 47.8 \\
 & v$\to$g & 0.0 & 0.0 & 0.0 & 0.0 & 0.0 & 0.0 & 0.0 & 0.0 \\
\midrule
\multirow{3}{*}{Gemma-2-9B~\citep{gemma2-2024}} & rot & 36.7 & 21.1 & 33.3 & 31.1 & 34.4 & 17.8 & 18.9 & 21.1 \\
 & rotChk & 7.8 & 0.0 & 50.0 & 26.7 & 27.8 & 10.0 & 7.8 & 73.3 \\
 & v$\to$g & 23.3 & 16.7 & 0.0 & 2.2 & 40.0 & 35.6 & 25.6 & 6.7 \\
\midrule
\multirow{3}{*}{Falcon3-10B~\citep{falcon3-2025}} & rot & 0.0 & 31.1 & 13.3 & 13.3 & 3.3 & 0.0 & 10.0 & 2.2 \\
 & rotChk & 18.9 & 25.6 & 74.4 & 36.7 & 47.8 & 40.0 & 28.9 & 58.9 \\
 & v$\to$g & 41.1 & 0.0 & 0.0 & 0.0 & 0.0 & 0.0 & 0.0 & 0.0 \\
\midrule
\rowcolor{rowbeige}
\multicolumn{10}{l}{\textbf{\textit{(B$'$) Mid-scale open-weight (27--32B)}}} \\
\multirow{3}{*}{Qwen2.5-32B~\citep{qwen2024qwen25}} & rot & 7.8 & 14.4 & 3.3 & 11.1 & 50.0 & 11.1 & 30.0 & 5.6 \\
 & rotChk & 10.0 & 36.7 & 1.1 & 13.3 & 16.7 & 27.8 & 27.8 & 38.9 \\
 & v$\to$g & 26.7 & 35.6 & 0.0 & 0.0 & 0.0 & 1.1 & 23.3 & 12.2 \\
\midrule
\multirow{3}{*}{Gemma-3-27B~\citep{gemma3-2025}} & rot & 36.7 & 23.3 & 43.3 & 46.7 & 33.3 & 36.7 & 0.0 & 6.7 \\
 & rotChk & 66.7 & 16.7 & 83.3 & 96.7 & 13.3 & 50.0 & 33.3 & 16.7 \\
 & v$\to$g & 3.3 & 40.0 & 0.0 & 23.3 & 10.0 & 0.0 & 20.0 & 33.3 \\
\midrule
\rowcolor{rowbeige}
\multicolumn{10}{l}{\textbf{\textit{(B$''$) Vision-language (text-only ablation)}}} \\
\multirow{3}{*}{Qwen2.5-VL-3B~\citep{qwen25vl-2025}} & rot & 0.0 & 0.0 & 0.0 & 43.0 & 10.7 & 0.0 & 0.0 & 0.0 \\
 & rotChk & 13.1 & 0.0 & 0.0 & 1.3 & 0.0 & 0.0 & 0.0 & 0.0 \\
 & v$\to$g & 12.6 & 0.0 & 0.0 & 0.0 & 28.5 & 0.0 & 0.0 & 0.0 \\
\bottomrule
\end{tabular}%
}
\caption{
\label{tab:app_l3_subtask_lang}
\textbf{L3 sub-task pass rates per language and model (direct).}
Three sub-tasks span the L3 task families:
\textbf{rot} (egocentric manipulation), \textbf{rotChk} (boolean probe),
\textbf{v$\to$g} (frame transformation).
}
\end{table}

%% file: tables/appendix_l5_graph.tex
%

\begin{table}[!htbp]
\centering
\scriptsize
\setlength{\tabcolsep}{3pt}
\renewcommand{\arraystretch}{1.05}
\resizebox{\columnwidth}{!}{%
\begin{tabular}{l l cccc}
\toprule
\textbf{Model} & \textbf{Sub-task} & \textbf{nF1\%} & \textbf{eF1\%} & \textbf{hallN} & \textbf{missN} \\
\midrule
\rowcolor{rowlavender}
\multicolumn{6}{l}{\textbf{\textit{(A) Frontier closed-source}}} \\
GPT-4o~\citep{openai2024gpt4o}            & static               & 100.0 & 97.1 & 0.00   & 0.00 \\
                  & dynamic              & 97.2  & 78.8 & 2.00   & 3.60 \\
                  & counterfactual       & 98.8  & 89.9 & 1.00   & 2.75 \\
                  & local                & 91.1  & 87.7 & 55.33  & 1.33 \\
\midrule
Gemini-2.5-Flash~\citep{google2025gemini25}  & static               & 44.9  & 30.4 & 26.50  & 75.25 \\
                  & dynamic              & 29.9  & 19.9 & 10.83  & 68.67 \\
                  & counterfactual       & 30.1  & 20.1 & 6.00   & 79.33 \\
                  & local                & 29.6  & 29.2 & 4.67   & 51.17 \\
\midrule
DeepSeek-V4-Flash~\citep{deepseek2025r1} & static               & 55.3  & 37.8 & 1.14   & 39.29 \\
                  & dynamic              & 55.3  & 44.8 & 1.71   & 38.71 \\
                  & counterfactual       & 53.4  & 45.8 & 1.43   & 44.00 \\
                  & local                & 50.0  & 49.0 & 13.29  & 33.29 \\
\midrule
Qwen3.5-Flash~\citep{qwen35flash2025} & static               & 100.0 & 100.0 & 0.00   & 0.00 \\
                  & dynamic              & 99.4  & 98.4  & 3.00   & 0.00 \\
                  & counterfactual       & 99.6  & 99.6  & 1.00   & 1.00 \\
                  & local                & 95.7  & 95.4  & 34.00  & 0.00 \\
\midrule
\rowcolor{rowbeige}
\multicolumn{6}{l}{\textbf{\textit{(B$^{\flat}$) Sub-cliff regime ($\leq$2B)}}} \\
SmolLM2-1.7B~\citep{allal2025smollm2}     & static               & 55.6  & 40.9 & 31.33  & 219.67 \\
                  & dynamic              & 35.8  & 18.3 & 52.00  & 335.00 \\
                  & counterfactual       & 43.8  & 24.8 & 30.67  & 324.33 \\
                  & local                & 47.8  & 34.4 & 65.67  & 365.00 \\
\midrule
\rowcolor{rowbeige}
\multicolumn{6}{l}{\textbf{\textit{(B) Open-weight (7--10B)}}} \\
Llama-3.1-8B~\citep{grattafiori2024llama3}      & static               & 83.6  & 30.9 & 1.00   & 6.86 \\
                  & dynamic              & 98.0  & 58.7 & 1.50   & 4.17 \\
                  & counterfactual       & 97.7  & 57.4 & 1.50   & 3.83 \\
                  & local                & 70.2  & 65.4 & 104.83 & 1.00 \\
\midrule
Qwen2.5-7B~\citep{qwen2024qwen25}        & static               & 81.7  & 41.3 & 4.00   & 30.86 \\
                  & dynamic              & 97.4  & 73.6 & 5.83   & 4.83 \\
                  & counterfactual       & 96.9  & 75.8 & 6.20   & 7.80 \\
                  & local                & 76.2  & 74.2 & 27.14  & 18.86 \\
\midrule
Mistral-7B~\citep{jiang2023mistral7b}        & static               & 78.9  & 31.8 & 3.67   & 14.50 \\
                  & dynamic              & 73.1  & 42.6 & 2.62   & 12.62 \\
                  & counterfactual       & 83.3  & 38.0 & 1.29   & 5.57 \\
                  & local                & 67.4  & 60.2 & 30.43  & 18.29 \\
\midrule
Gemma-2-9B~\citep{gemma2-2024}        & static               & 99.0  & 88.2 & 1.20   & 2.60 \\
                  & dynamic              & 98.8  & 61.9 & 2.00   & 1.80 \\
                  & counterfactual       & 97.9  & 74.4 & 1.33   & 3.50 \\
                  & local                & 63.2  & 57.2 & 38.25  & 4.00 \\
\midrule
Falcon3-10B~\citep{falcon3-2025}       & static               & 97.2  & 74.2 & 3.40   & 4.00 \\
                  & dynamic              & 68.5  & 59.6 & 3.14   & 9.14 \\
                  & counterfactual       & 81.7  & 71.4 & 2.00   & 7.86 \\
                  & local                & 59.2  & 59.2 & 39.43  & 6.14 \\
\midrule
\rowcolor{rowbeige}
\multicolumn{6}{l}{\textbf{\textit{(B$'$) Mid-scale open-weight (27--32B)}}} \\
Qwen2.5-32B~\citep{qwen2024qwen25}     & static               & 99.6  & 99.6 & 0.00   & 2.00 \\
                  & dynamic              & 98.4  & 90.4 & 4.51   & 1.18 \\
                  & counterfactual       & 99.4  & 97.0 & 1.42   & 0.53 \\
                  & local                & 89.8  & 88.6 & 50.00  & 8.67 \\
\midrule
Gemma-3-27B~\citep{gemma3-2025}        & static               & 100.0 & 72.2 & 0.00   & 0.00 \\
                  & dynamic              & 99.0  & 88.2 & 1.00   & 1.50 \\
                  & counterfactual       & 98.3  & 70.8 & 0.00   & 5.00 \\
                  & local                & 100.0 & 80.0 & 0.00   & 0.00 \\
\bottomrule
\end{tabular}%
}
\caption{
\label{tab:app_l5_graph_diagnostics}
\textbf{L5 graph diagnostics per (model, sub-task) in English, direct.}
\textbf{nF1}/\textbf{eF1}: token-level node/edge F1 against the ProcTHOR
ground-truth scene graph. \textbf{hallN}/\textbf{missN}: average count of
hallucinated and missing nodes per response. The dissociation between node
identification and relation extraction (highest on \texttt{static} sub-task)
supports F4.
}
\end{table}

%% file: tables/appendix_json_validity.tex
\begin{table}[!htbp]
\centering
\small
\setlength{\tabcolsep}{4pt}
\renewcommand{\arraystretch}{1.1}
\begin{tabular}{lrrr}
\toprule
\textbf{Model} & \textbf{Direct} & \textbf{Reasoning} & \textbf{$\Delta$} \\
\midrule
\rowcolor{rowlavender}
\multicolumn{4}{l}{\textbf{\textit{(A) Frontier closed-source}}} \\
GPT-4o                  & 83.3\% & 66.4\% & $-16.9$ \\
Gemini-2.5-Flash        & 36.0\% & 31.3\% & $-4.7$ \\
DeepSeek-V4-Flash       & 58.0\% & 63.1\% & $+5.2$ \\
Qwen3.5-Flash           & 82.8\% & 60.4\% & $-22.4$ \\
\midrule
\rowcolor{rowbeige}
\multicolumn{4}{l}{\textbf{\textit{(B$^{\flat}$) Sub-cliff regime ($\leq$2B)}}} \\
SmolLM2-1.7B            & 58.8\% & 61.5\% & $+2.7$ \\
\midrule
\rowcolor{rowbeige}
\multicolumn{4}{l}{\textbf{\textit{(B) Open-weight (7--10B)}}} \\
Llama-3.1-8B            & 80.6\% & 46.7\% & $-33.9$ \\
Qwen2.5-7B              & 79.9\% & 72.2\% & $-7.6$ \\
Mistral-7B-v0.3         & 70.4\% & \phantom{0}7.3\% & $-63.0$ \\
Gemma-2-9B              & 80.4\% & 50.6\% & $-29.9$ \\
Falcon3-10B             & 72.2\% & 68.5\% & $-3.7$ \\
\midrule
\rowcolor{rowbeige}
\multicolumn{4}{l}{\textbf{\textit{(B$'$) Mid-scale open-weight (27--32B)}}} \\
Qwen2.5-32B             & 83.3\% & 64.6\% & $-18.7$ \\
Gemma-3-27B-it          & 100.0\% & 80.4\% & $-19.6$ \\
\midrule
\rowcolor{rowbeige}
\multicolumn{4}{l}{\textbf{\textit{(B$''$) Vision-language (text-only ablation)}}} \\
Qwen2.5-VL-3B           & 80.9\% & 80.3\% & $-0.5$ \\
\bottomrule
\end{tabular}
\caption{\label{tab:app_json_validity}
\textbf{L5 JSON validity rate by prompt regime.}
Percentage of L5 cells producing parseable JSON with the required \texttt{nodes}/\texttt{edges}
schema, pooled across all eight natural languages, all three description modes,
and all five L5 sub-tasks. Ten of eleven models show degraded JSON validity
under reasoning prompts; only DeepSeek-V4-Flash improves. Mistral-7B
collapses from 70.4\% to 7.3\% (the most extreme case), consistent with
the F3 mechanism that reasoning preambles compete with strict JSON token
budgets. The single positive case (DeepSeek) is also the model with the
largest positive L3 reasoning delta (+32.4pp), suggesting its decoding stack
allocates output budget differently from the other systems.
}
\end{table}

%% file: tables/appendix_anchor_preference.tex
\begin{table}[!htbp]
\centering
\small
\setlength{\tabcolsep}{4pt}
\renewcommand{\arraystretch}{1.1}
\begin{tabular}{lrrrr}
\toprule
& \multicolumn{2}{c}{\textbf{LTR (en/es/de)}} & \multicolumn{2}{c}{\textbf{RTL (ar)}} \\
\cmidrule(lr){2-3} \cmidrule(lr){4-5}
\textbf{Model} & \textbf{Source} & \textbf{Target} & \textbf{Source} & \textbf{Target} \\
\midrule
\rowcolor{rowlavender}
\multicolumn{5}{l}{\textbf{\textit{(A) Frontier closed-source}}} \\
DeepSeek-V4-Flash       & 2.0\% & 30.9\% & 0.0\% & 0.0\% \\
Qwen3.5-Flash           & 13.2\% & 20.1\% & 0.0\% & 0.0\% \\
\midrule
\rowcolor{rowbeige}
\multicolumn{5}{l}{\textbf{\textit{(B$^{\flat}$) Sub-cliff regime ($\leq$2B)}}} \\
SmolLM2-1.7B            & 17.6\% & 19.0\% & 8.5\% & 3.0\% \\
\midrule
\rowcolor{rowbeige}
\multicolumn{5}{l}{\textbf{\textit{(B) Open-weight (7--10B)}}} \\
Llama-3.1-8B            & 0.7\% & 32.6\% & 1.7\% & 0.6\% \\
Qwen2.5-7B              & 1.1\% & 32.2\% & 0.0\% & 0.0\% \\
Mistral-7B-v0.3         & 13.5\% & 25.7\% & 16.7\% & 2.8\% \\
Gemma-2-9B              & 6.1\% & 27.4\% & 0.0\% & 0.0\% \\
Falcon3-10B             & 3.5\% & 28.0\% & 0.0\% & 0.0\% \\
\midrule
\rowcolor{rowbeige}
\multicolumn{5}{l}{\textbf{\textit{(B$'$) Mid-scale open-weight (27--32B)}}} \\
Qwen2.5-32B             & 3.3\% & 30.0\% & 0.0\% & 1.7\% \\
Gemma-3-27B-it          & 3.9\% & 29.4\% & 0.0\% & 0.0\% \\
\midrule
\rowcolor{rowbeige}
\multicolumn{5}{l}{\textbf{\textit{(B$''$) Vision-language (text-only ablation)}}} \\
Qwen2.5-VL-3B           & 0.6\% & 34.5\% & 0.0\% & 0.0\% \\
\bottomrule
\end{tabular}
\caption{\label{tab:app_anchor_preference}
\textbf{RTL anchor-preference probe.} Per-model rates at which the model
selects the source vs.\ target as the reference anchor when describing a
spatial relation, on the \texttt{rtlAnchorPreference} task. In LTR
languages (en/es/de averaged), all evaluated models prefer the
\emph{target} anchor by a margin of 19--34pp---a consistent cross-model
pattern in the LTR setting. The Arabic and Thai cells fall predominantly
into the \emph{unknown} bucket, reflecting the scope of the current
anchor-extraction protocol on non-Latin scripts; cross-language anchor
preference under a multilingually-calibrated extractor is a natural
follow-up extension.}
\end{table}